\documentclass{article}

\usepackage{dblfloatfix}

\usepackage{microtype}
\usepackage{graphicx}
\usepackage{subcaption}
\usepackage{booktabs} 
\usepackage{enumitem}
\usepackage{tabularx}
\usepackage{array}
\usepackage{makecell}

\usepackage{hyperref}

\usepackage[preprint]{icml2026}

\usepackage{amsmath}
\usepackage{amssymb}
\usepackage{mathtools}
\usepackage{amsthm}

\usepackage[capitalize,noabbrev]{cleveref}

\theoremstyle{plain}

\theoremstyle{definition}

\theoremstyle{remark}

\usepackage[textsize=tiny]{todonotes}

\newcommand{\gbt}{\textsc{GBT}}

\icmltitlerunning{Traversal-as-Policy: Log-Distilled Gated Behavior Trees as Externalized, Verifiable Policies for Safe, Robust, and Efficient Agents}

\begin{document}

\twocolumn[
  \icmltitle{Traversal-as-Policy: Log-Distilled Gated Behavior Trees as Externalized, Verifiable Policies for Safe, Robust, and Efficient Agents}



  \icmlsetsymbol{equal}{*}

  \begin{icmlauthorlist}
    \icmlauthor{Peiran Li}{a,b}   
    \icmlauthor{Jiashuo Sun}{d}
    \icmlauthor{Fangzhou Lin}{a,c}
    \icmlauthor{Shuo Xing}{a} \\ 
    \icmlauthor{Tianfu Fu}{e}
    \icmlauthor{Suofei Feng}{f} 
    \icmlauthor{Chaoqun Ni}{b}
    \icmlauthor{Zhengzhong Tu}{a}    
  \end{icmlauthorlist}

  \icmlaffiliation{a}{Texas A\&M University}  
  \icmlaffiliation{b}{University of Wisconsin-Madison}
  \icmlaffiliation{c}{Worcester Polytechnic Institute}  
  \icmlaffiliation{d}{University of Illinois Urbana-Champaign}    
  \icmlaffiliation{e}{xAI} \icmlaffiliation{f}{Meta}     

  \icmlcorrespondingauthor{Peiran Li}{lipeiran@tamu.edu}
  \icmlcorrespondingauthor{Zhengzhong Tu}{tzz@tamu.edu}

  \icmlkeywords{Machine Learning, ICML}

  \vskip 0.3in

{
\begin{center}
\vspace{-20px}
    \centering
    \captionsetup{type=figure}
    \includegraphics[width=\textwidth]
    {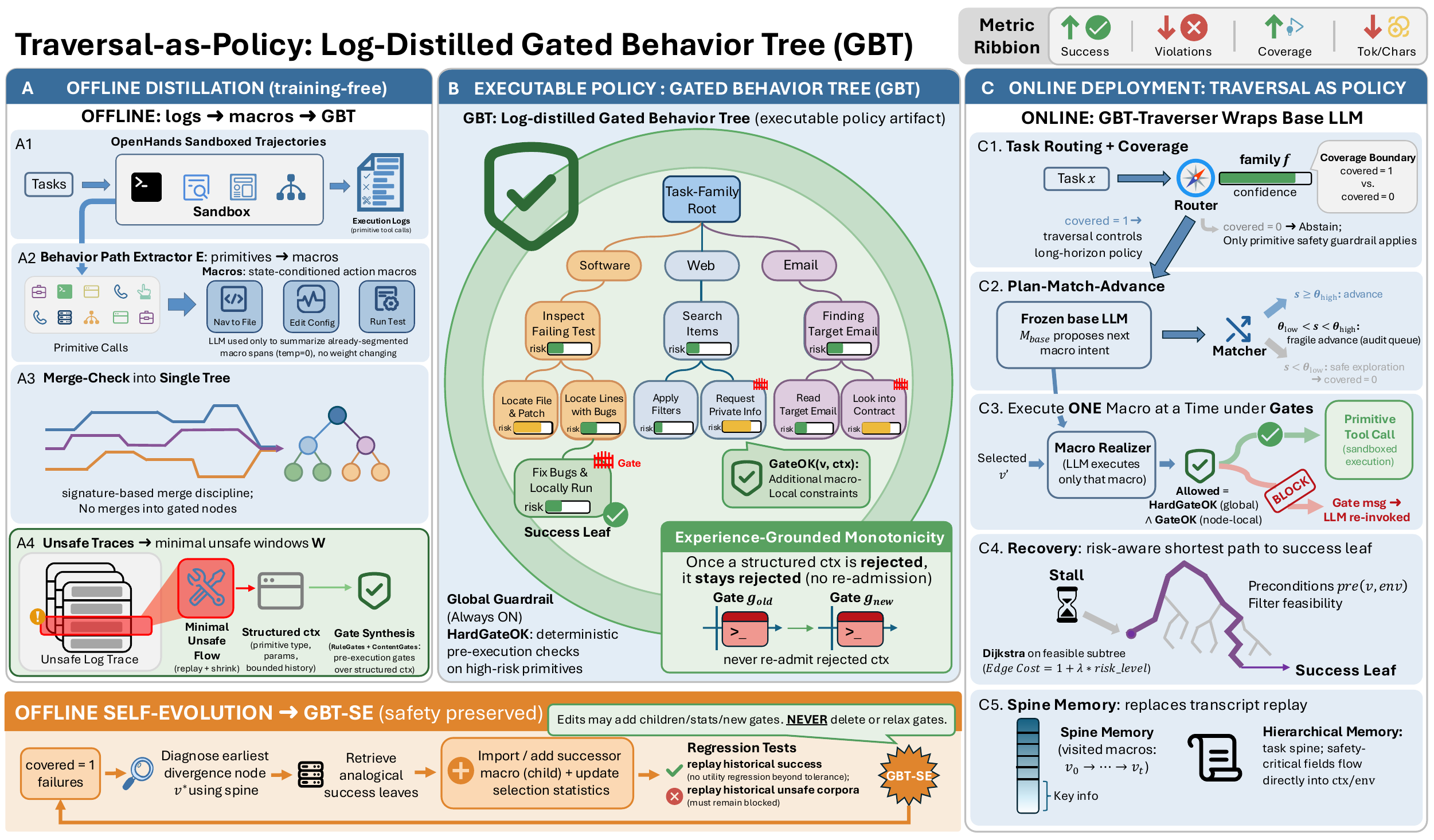}
    \captionof{figure}{\textbf{Overview of Traversal-as-Policy.} Gated Behavior Tree (GBT) distills offline logs into a gated tree artifact. Online traversal wraps a base LLM, executing macros under deterministic gates and using risk-aware recovery. Self-evolution safely evolves the policy under strict regression invariants.
}
    \label{fig:teaser}
\end{center}%
\vskip 0.1in
}

]



\printAffiliationsAndNotice{}  

\begin{abstract}
Autonomous LLM agents fail because long-horizon policy remains implicit in model weights and transcripts, while safety is retrofitted post hoc. We propose \emph{Traversal-as-Policy}: distill sandboxed OpenHands execution logs into a single executable \textbf{Gated Behavior Tree} (\gbt{}) and treat tree traversal---rather than unconstrained generation---as the control policy whenever a task is in coverage. Each node encodes a \emph{state-conditioned action macro} mined and merge-checked from successful trajectories; macros implicated by unsafe traces attach deterministic \emph{pre-execution gates} over structured tool context and bounded history, updated under experience-grounded monotonicity so previously rejected unsafe contexts cannot be re-admitted. At runtime, a lightweight traverser matches the base model's intent to child macros, executes one macro at a time under global and node-local gating, and when stalled performs risk-aware shortest-path recovery to a feasible success leaf; the visited path forms a compact spine memory that replaces transcript replay. Evaluated in a unified OpenHands sandbox on \textit{15+} software, web, reasoning, and safety/security benchmarks, \gbt{} improves success while driving violations toward zero and reducing cost. On SWE-bench Verified (Protocol A, 500 issues), \gbt{}-SE raises success from 34.6\% to 73.6\%, reduces violations from 2.8\% to 0.2\%, and cuts token/character usage from 208k/820k to 126k/490k; with the same distilled tree, 8B executors more than double success on SWE-bench Verified (14.0\%\,$\rightarrow$\,58.8\%) and WebArena (9.1\%\,$\rightarrow$\,37.3\%).

\end{abstract}


\section{Introduction}

Autonomous language-model agents are increasingly asked to \emph{act}: patch repositories, navigate websites, run commands, and chain tools over long horizons. Yet most deployments still execute an \emph{implicit} policy -- a brittle mixture of model weights, prompt templates, and ad-hoc transcripts -- while safety is retrofitted post hoc. The policy remains buried in weights and logs, making agents hard to debug, difficult to certify, and expensive to improve.

Reasoning and safety advances have not produced an explicit, reusable control object. Deliberative search and reflection-style memories (Tree-of-Thoughts; Reflexion; Meta-Policy Reflexion) still decide and remember through free-form generation, treating logs as transient context rather than compiling them into an executable artifact \citep{treeofthought,reflexion,mpr}. Guardrailing attaches runtime validators or ``guardian'' agents \citep{guardagent,aworld,policyasprompt}, but their safety knowledge is typically human-specified (rules, prompts, or code), unscalable, and blind to long-tail, context-dependent failures that only appear in operational traces.

We argue what is missing is a \emph{first-class, model-external policy artifact} distilled from execution that unifies: (i) robust long-horizon control, (ii) deterministic safety \emph{before} high-risk actions, and (iii) a compact state representation that replaces transcript replay. This motivates a concrete question: \emph{Can we distill massive execution logs into a single executable artifact that simultaneously governs behavior, enforces safety, and compresses memory -- avoiding weight updates?}

We answer yes with \emph{Traversal-as-Policy}. From sandboxed OpenHands trajectories \citep{wang2024openhands}, we distill a single executable \textbf{Gated Behavior Tree} (\gbt{}), and treat \emph{tree traversal} -- rather than unconstrained generation -- as the control policy whenever a task is in coverage. Each node is a \emph{state-conditioned action macro}: a contiguous segment of primitive tool calls that stays within a local sub-action region and realizes a coherent intent. Successful trajectories contribute macro paths that are merge-checked into a single tree under a signature-based discipline designed to prevent semantic aliasing where safety matters. Coverage is explicit: when semantic matching is confident, \gbt{} governs long-horizon control; otherwise the traverser abstains and the episode is labeled \texttt{covered}=0.

Safety enters \gbt{} through \emph{experience-grounded pre-execution gates}. We designate high-risk primitives (writes/deletes, process spawns, network sends, sensitive reads). For each attempted high-risk primitive we construct a structured context \texttt{ctx} directly from sandbox state (primitive type, parameters such as file paths or domains, process metadata, and a bounded history). Gates are deterministic checks over these structured fields -- they do not consult LLM summaries -- so safety cannot be bypassed by prompt hacking or summarization choices. Unsafe traces are deterministically replayed and shrunk to minimal violating windows; the extracted contexts synthesize global or node-local gates. Gates update under \emph{experience-grounded monotonicity}: once a structured context is rejected, it stays rejected by all future gate libraries, preventing silent safety regression.

At runtime, a lightweight \textbf{GBT-Traverser} makes \gbt{} operational. The base model proposes the next step at the level of intent; the traverser matches this proposal to a child macro and advances only along that transition, executing one macro at a time under both global and node-local gating. Traversal arrests long-horizon drift by constraining choices to successors grounded in prior successes; when stalled or repeatedly blocked, the traverser performs risk-aware shortest-path recovery to a feasible success leaf. \gbt{} also serves as hierarchical memory: the traverser maintains a compact \emph{task spine} (visited macros) as persistent long-term state, replacing transcript replay; safety-critical fields bypass summarization and flow directly into \texttt{ctx}. Offline, a failure-driven self-evolution loop upgrades \gbt{}-Basic to \gbt{}-SE by locally repairing \texttt{covered}=1 failures via analogical successes and updated selection statistics, while preserving safety by forbidding gate relaxation and regression-testing historical successes and historical unsafe corpora.

Experiments evaluate whether \gbt{} behaves as a first-class external policy artifact under strictly controlled OpenHands execution \citep{wang2024openhands}. On SWE-bench Verified~\cite{jimenez2023swe} (Protocol A, 500 issues), \gbt{}-SE raises success from 34.6\% to 73.6\% at 86.0\% coverage, reduces violations from 2.8\% to 0.2\%, eliminates unsafe success (1.2\%$\rightarrow$0.0\%), and cuts Tok/Chars (thousands) from 208/820 to 126/490; global guardrail alone yields 38.8\% success with 0.8\% violations. On WebArena~\cite{zhou2023webarena} (Protocol A, 812 tasks), \gbt{}-SE raises success from 19.7\% to 66.9\% at 78.0\% coverage, reduces violations from 3.4\% to 0.2\%, eliminates unsafe success (1.0\%$\rightarrow$0.0\%), and cuts Tok/Chars from 94/360 to 52/205; global guardrail alone leaves success essentially unchanged (19.3\%). On GPQA~\cite{rein2024gpqa} (Protocol A, 448; browsing disabled), \gbt{}-SE raises accuracy from 58.7\% to 87.3\% at 73.0\% coverage while keeping violations at 0.2\% and unsafe success at 0.0\%, and reduces Tok/Chars from 22/86 to 15/58. After distillation, the same tree enables small executors: \texttt{Qwen3-VL-8B-Thinking}~\cite{bai2025qwen3vltechnicalreport} rises from 14.0\% to 58.8\% on SWE-bench Verified and from 9.1\% to 37.3\% on WebArena, demonstrating decoupling of offline reasoning from online execution.

\textbf{Contributions:}
\vspace{-.2em}
\begin{itemize}[leftmargin=*, itemsep=0.1em, topsep=0.1em, parsep=0pt, partopsep=0pt]
    \item \textbf{Traversal-as-Policy:} we externalize long-horizon agent control as executable tree traversal in a log-distilled \textbf{Gated Behavior Tree} (\gbt{}), yielding a persistent policy artifact independent of model weights.
    \item \textbf{Experience-grounded deterministic safety:} we compile unsafe traces into \emph{pre-execution} gates over structured contexts and update them under monotonicity so logged unsafe contexts cannot be re-admitted.
    \item \textbf{Unified evidence:} on \textit{15+} OpenHands benchmarks, \gbt{} simultaneously improves success, reduces violations and unsafe success toward zero, cuts inference cost, and enables strong execution by small models.
\end{itemize}

\section{Gated Behavior Tree (\gbt)}
\label{sec:method}

We instantiate \emph{Traversal-as-Policy} by distilling sandboxed execution logs into a single \emph{Gated Behavior Tree} (\gbt{}), then treating \emph{tree traversal}---rather than unconstrained generation---as the agent's long-horizon control policy whenever the task is within coverage. The pipeline is \emph{training-free}: we never update model weights. Frozen LLMs (temperature $0$) are used only offline to summarize chosen macro segments for semantic matching, synthesize deterministic predicates, and diagnose failures; their outputs are compiled into inspectable artifacts (tree structure, gate predicates, and preconditions). Throughout all phases (data collection, offline distillation, and online deployment), the same primitive-level guardrail vets each high-risk tool call.


\textbf{Pipeline (3-line overview):}
\vspace{-0.2em}
\begin{itemize}[leftmargin=*, itemsep=0.1em, topsep=0.1em, parsep=0pt, partopsep=0pt]
    \item \textbf{Offline:} logs $\rightarrow$ primitives $\rightarrow$ macros $\rightarrow$ merge into \emph{GBT-Basic} + attach gates from unsafe windows.
    \item \textbf{Offline:} covered failures $\rightarrow$ local repair via analogical successes $\rightarrow$ \emph{GBT-SE} (safety preserved under monotone invariants).
    \item \textbf{Online:} traversal policy + primitive gating + recovery search + spine memory (coverage-scoped claims).
\end{itemize}

\textbf{What is a macro?}
A \emph{state-conditioned action macro} is a of primitive tool calls that (i) stays within a local sub-action region of the environment and (ii) realizes a coherent intent. For example, in software tasks a macro might be ``inspect failing test and open the referenced file,'' while in web tasks it might be ``fill remaining form fields and submit.'' The key separation is \emph{macro-level control} (what step to do next) versus \emph{primitive-level execution} (the concrete tool calls), with safety enforced at the primitive level.

\textbf{Three coupled safety/control objects.}
Safety and control are coupled through: (i) a sandbox-level \emph{normative} safety specification $\mathcal{S}_{\text{spec}}$ defined by monitors and benchmark checkers; (ii) an \emph{executable} bounded-history approximation $\mathcal{S}_{\text{sys}}(t)$ using pre-execution gates over structured contexts; and (iii) a log-distilled tree whose nodes encode reusable macros and may carry \emph{node-local} gates. When \gbt{} is in coverage, the executed long-horizon behavior is the realized traversal path, while \emph{every} high-risk primitive is still mediated by deterministic, pre-execution gate checks.

\subsection{Safety Objects and Experience-Grounded Gates}
\label{sec:safety}
\vspace{-.2em}
\textbf{Sandboxed trajectories and structured contexts.}
A trajectory $\tau$ is labeled \texttt{unsafe} iff at least one primitive tool call violates $\mathcal{S}_{\text{spec}}$ in the OpenHands sandbox~\citep{wang2024openhands}. For high-risk primitives we construct a structured context $\texttt{ctx}\in\mathcal{C}$ directly from sandbox state: primitive type, parameters (paths, domains, process metadata, etc.), and a bounded history of recent high-risk primitives along the same macro. \emph{All safety decisions depend only on these structured fields.} LLM summaries are never read by gates or preconditions (details and examples in App.~\ref{app:guardrail}).

\textbf{Gate interface and executable subset.}
Each gate has the common interface
$$
g:\ \texttt{ctx} \mapsto (\texttt{ok}, \texttt{msg}) \in \{\texttt{true},\texttt{false}\}\times\texttt{String}.
$$
Let $\mathcal{G}(t)=\mathcal{G}_{\text{global}}(t)\cup\mathcal{G}_{\text{node}}(t)$ denote global plus node-local gates at time $t$. The executable safety subset is
\[
\mathcal{S}_{\text{sys}}(t)
=
\left\{\texttt{ctx}\in\mathcal{C}:\exists g\in\mathcal{G}(t),\ g(\texttt{ctx}).\texttt{ok}=\texttt{false}\right\},
\]
a conservative approximation to the portion of $\mathcal{S}_{\text{spec}}$ expressible over bounded-history structured contexts. \emph{Consequently, no summarization or prompting choice can bypass safety:} summaries can influence how a macro is realized, but cannot change $\texttt{ctx}$ nor the gate outcomes computed from it.

\textbf{Experience-grounded monotonicity.}
For every gate $g$ we maintain an unsafe corpus $\mathcal{D}_{\text{unsafe}}(g)$ (contexts extracted from observed violations) and a benign corpus $\mathcal{D}_{\text{benign}}(g)$ (representative safe contexts for the same primitive family). Any update from $g^{\text{old}}$ to $g^{\text{new}}$ must satisfy:
\[
\begin{aligned}
\forall\, \texttt{ctx}\in\mathcal{D}_{\text{unsafe}}(g):\quad
&g^{\text{old}}(\texttt{ctx}).\texttt{ok}=\texttt{false}\\
&\Rightarrow\;
g^{\text{new}}(\texttt{ctx}).\texttt{ok}=\texttt{false}.
\end{aligned}
\]
and control false positives on benign contexts:
\[
\frac{\left|\left\{\texttt{ctx}\in\mathcal{D}_{\text{benign}}(g): g^{\text{new}}(\texttt{ctx}).\texttt{ok}=\texttt{false}\right\}\right|}{|\mathcal{D}_{\text{benign}}(g)|}
\le \epsilon_{\text{benign}}.
\]
We give concrete corpora construction and gate update procedures in App.~\ref{app:guardrail}.

\textbf{Design invariant 1 (Experience-grounded monotonicity).}
Once a structured context is recorded as unsafe and rejected by any gate, it stays rejected by all future gate libraries.\\
\emph{Consequence:} the system can only \emph{expand} $\mathcal{S}_{\text{sys}}(t)$ on observed violations; it cannot re-admit logged unsafe contexts.

\subsection{Global Guardrail}
\label{sec:guardrail}
\vspace{-.2em}
We designate a set of high-risk primitives $\mathcal{T}_{\text{hard}}$ (writes/deletes, process spawns, network sends, and reads of resources matching sensitive patterns). For any candidate primitive $a\in\mathcal{T}_{\text{hard}}$ with structured context $\texttt{ctx}$, the global guardrail evaluates a fixed but extensible library $\mathcal{G}_{\text{global}}$ consisting of:
(i) \emph{RuleGates}, deterministic code predicates over structured fields and bounded history; and
(ii) \emph{ContentGates}, deterministic calls (temperature $0$) to a fixed guard classifier over payload-bearing actions, post-processed into a Boolean flag and message.
The global decision is
\[
\texttt{HardGateOK}(a,\texttt{ctx})
=
\bigwedge_{g\in\mathcal{G}_{\text{global}}} g(\texttt{ctx}).\texttt{ok}.
\]
Unsafe trajectories that still occur under the current guardrail expose violations of $\mathcal{S}_{\text{spec}}$ not yet captured by $\mathcal{S}_{\text{sys}}(t)$. We deterministically replay and shrink such traces to minimal violating primitive windows, record their structured contexts into unsafe corpora, and synthesize new global or node-local gates under Design invariant~1 (App.~\ref{app:guardrail}).

\subsection{From Logs to \texorpdfstring{GBT-Basic}{GBT-Basic}}
\label{sec:gbt-basic}
\vspace{-.2em}
\textbf{Macro abstraction (behavior paths).}
A frozen \emph{Behavior Path Extractor} $E$ maps a trajectory $\tau$ to a short macro path
\[
p=(v_0,\dots,v_K)=E(\tau),
\]
where each macro $v_k$ is a state-conditioned action macro. Macro boundaries are anchored by observable environment deltas (file diffs, domain changes, process starts) and tool-usage heuristics; LLM calls, when used, only summarize already-segmented macro spans into short descriptions for matching. Because traversal assumes macros have stable semantics, we explicitly test abstraction stability on unsafe or $\mathcal{T}_{\text{hard}}$-touching traces and exclude abstract-unstable trajectories from tree construction and gate derivation (App.~\ref{app:behavior-extractor}).

\textbf{Unsafe windows $\rightarrow$ node-local gates.}
For an unsafe trajectory, we shrink it (via sandbox replay) to a minimal unsafe primitive window $W$, map it to a minimal covering macro subsequence $U$ in $E(\tau)$, and treat each macro $v\in U$ as participating in the unsafe pattern. We attach to such nodes a small set of node-local gates $\mathcal{G}(v)$ with the same interface $g:\texttt{ctx}\mapsto(\texttt{ok},\texttt{msg})$, synthesized from observed violating contexts and constrained by the same monotonicity and benign-regression rules as global gates (App.~\ref{app:guardrail}).

\textbf{Tree construction and merge discipline.}
\gbt{} is a single rooted tree whose first layer consists of task-family roots that reduce branching; a frozen router assigns each task to a family when confident (App.~\ref{app:family-classifier}). Each macro node $v$ stores: a natural-language \texttt{description} for semantic matching, coarse tags, a discrete risk level, and a behavior signature $\sigma(v)=(\sigma_{\text{disc}}(v),\sigma_{\text{cont}}(v))$ computed purely from logs. When inserting a successful path, we reuse an existing child only when signatures and descriptions agree (exact match on $\sigma_{\text{disc}}$, cosine similarity thresholds on $\sigma_{\text{cont}}$ and embedded descriptions). To prevent semantic aliasing where safety matters most, merges into nodes that already carry node-local gates are disallowed. High-traffic ungated nodes are audited for aliasing and split when needed; acyclicity is enforced as a hard invariant (App.~\ref{app:tree-construction}). The resulting \emph{GBT-Basic} encodes reusable macros, explicit success leaves, and log-grounded node-local gates, while remaining coupled to the global guardrail that protects all high-risk primitives everywhere.

\textbf{Design invariant 2 (Tree edits do not weaken enforcement).}
Gates depend only on structured $\texttt{ctx}$ and bounded history, not on node descriptions or topology; tree edits never delete/relax gates, and merges into gated nodes are disallowed.\\
\emph{Consequence:} changing the tree can alter behavior availability and selection frequency, but cannot turn any previously rejected unsafe context into an allowed one.

\subsection{Self-Evolution Under Safety Invariants: \texorpdfstring{GBT-SE}{GBT-SE}}
\label{sec:gbt-se}
\vspace{-.2em}
Many failures do not violate $\mathcal{S}_{\text{spec}}$; they are dominated by control and context drift. We run a failure-driven self-evolution loop that improves behavior \emph{locally} while preserving all accumulated safety invariants.

\textbf{Coverage predicate.}
Self-evolution operates only on episodes where \gbt{} genuinely serves as the policy skeleton (\texttt{covered}$=1$): the task is routed to a single family with high confidence, macro transitions match existing children above a minimum similarity threshold, and traversal never invokes safe exploration (full definition in App.~\ref{app:self-evolution}). Outside coverage (\texttt{covered}$=0$), we do \emph{not} claim long-horizon control from \gbt{}; we only claim primitive-level safety from the guardrail.

\enlargethispage{1\baselineskip} 
\textbf{Local repair via analogical successes.}
Given a failed covered path $p^{\text{fail}}=(v_0,\dots,v_K)$, a frozen reasoning model diagnoses failure on the \emph{task spine} (macro descriptions plus coarse environment summaries), backtracking to the earliest node $v^\star$ where choosing a different successor could plausibly avert failure. To repair $v^\star$, we retrieve analogical success leaves by embedding task descriptions, identify a matched success path whose corresponding node best aligns with $v^\star$, and import its successor macro. If an existing child of $v^\star$ already matches sufficiently, we reuse it; otherwise we add a new child whose description, behavior signature, and risk metadata mirror the imported successor. Each child $u$ of $v^\star$ maintains a selection score
\begin{equation}
\begin{aligned}
\texttt{score}(u;\texttt{ctx})
&=
\alpha\cdot \mathrm{sim}_\text{text}\big(\text{proposal},\text{desc}(u)\big)\\
&\quad+\;
\beta\cdot \mathrm{success\_rate}\big(u;\mathrm{cluster}(\texttt{ctx})\big),
\end{aligned}
\label{eq:score}
\end{equation}
combining semantic match to the base model’s proposal with empirical success statistics for similar structured contexts (definitions and maintenance in App.~\ref{app:self-evolution}).

\textbf{Safety preservation by construction.}
Self-evolution may add children, update local selection statistics, and add new gates for newly observed unsafe patterns; it may \emph{not} delete or weaken any gate. Any new child inherits the full node-local gate set of its parent before additional gates are allowed. Proposed edits are regression-tested by replaying historical successes through $v^\star$ (requiring success not to degrade beyond a small tolerance) and replaying historical unsafe episodes that previously triggered gates (verifying they remain blocked), as detailed in App.~\ref{app:self-evolution}.

\textbf{Design invariant 3 (Safety under self-evolution).}
Under the allowed edits and regression tests, self-evolution cannot re-admit previously rejected unsafe contexts.\\
\emph{Consequence:} \emph{GBT-SE} expands coverage/robustness while preserving Design invariants~1--2.

\subsection{Online Traversal-as-Policy}
\label{sec:online}
\vspace{-.2em}
At deployment, a \emph{GBT-Traverser} runs alongside a base model $M_{\text{base}}$. It routes tasks to a family subtree when confident, constrains macro choices by traversal, enforces primitive-level gating, triggers recovery when stalled, and uses the tree as hierarchical memory. The global guardrail remains active on every $a\in\mathcal{T}_{\text{hard}}$ regardless of coverage.

\textbf{Routing, coverage boundary, and abstention.}
Given a task description $x$, a frozen router predicts $p(f\mid x)$ over task families (App.~\ref{app:family-classifier}). If $\max_f p(f\mid x)$ is below a threshold, the traverser \emph{abstains} from traversal control and the episode is \texttt{covered}$=0$. Otherwise, traversal proceeds within the selected family subtree. We make the claim boundary explicit:
\[
\begin{aligned}
\texttt{covered}=1:&\quad
\text{\parbox[t]{0.6\linewidth}{\gbt\ determines long-horizon policy by traversal.}}\\
\texttt{covered}=0:&\quad
\text{\parbox[t]{0.6\linewidth}{we only claim primitive-level safety from the guardrail.}}
\end{aligned}
\]

\textbf{Plan--match--advance and safe exploration.}
At node $v_t$, $M_{\text{base}}$ proposes the next macro as a short description. The traverser embeds this proposal and matches it against children of $v_t$, yielding similarity $s$. Two thresholds $\theta_{\text{high}}>\theta_{\text{low}}$ define three regimes: high-confidence advance ($s\ge\theta_{\text{high}}$), fragile advance ($\theta_{\text{low}}\le s<\theta_{\text{high}}$, queued for offline inspection), and \emph{safe exploration} ($s<\theta_{\text{low}}$), where traversal temporarily abstains from steering long-horizon control around $v_t$ for a small budget. Any episode invoking safe exploration is marked \texttt{covered}$=0$.

\textbf{Primitive-level enforcement during macro realization.}
Once a child $v'$ is selected (by traversal or recovery), the traverser instructs $M_{\text{base}}$ to realize only this macro. For each executed high-risk primitive $a\in\mathcal{T}_{\text{hard}}$ with structured context $\texttt{ctx}_{\text{tool}}$, we require
\begingroup
\setlength{\abovedisplayskip}{4pt}
\setlength{\belowdisplayskip}{4pt}
\setlength{\abovedisplayshortskip}{2pt}
\setlength{\belowdisplayshortskip}{2pt}
\[
\begin{aligned}
\texttt{GateOK}(v',\texttt{ctx})
&=
\bigwedge_{g\in\mathcal{G}(v')} g(\texttt{ctx}).\texttt{ok},\\
\texttt{Allowed}(a,\texttt{ctx}_{\text{tool}},v')
&=
\texttt{HardGateOK}(a,\texttt{ctx}_{\text{tool}})\\
&\quad\land\;
\texttt{GateOK}(v',\texttt{ctx}_{\text{tool}}).
\end{aligned}
\]
\endgroup

If \texttt{Allowed} is false, the primitive is blocked and $M_{\text{base}}$ is reinvoked with gate messages. Because gates read only structured $\texttt{ctx}$, summarization cannot override or route around these checks.

\textbf{Recovery via risk-aware shortest paths.}
When the agent stalls or repeatedly trips gates, control inverts: \gbt{} proposes recovery macros and $M_{\text{base}}$ executes them. We retrieve candidate success leaves similar to the task and filter them by a coarse environment signature match. Each node carries a deterministic precondition $\texttt{pre}(v,\texttt{env})$ over a structured environment summary; preconditions are conservative feasibility filters and never override gates (App.~\ref{app:recovery}). Restricted to feasible nodes and matched leaves, we run Dijkstra on the family subtree with edge cost
{%
\setlength{\abovedisplayskip}{6pt}
\setlength{\belowdisplayskip}{6pt}
\setlength{\abovedisplayshortskip}{4pt}
\setlength{\belowdisplayshortskip}{4pt}
\begin{equation}
c(v\to u)=1+\lambda\cdot \mathrm{risk\_level}(u),
\label{eq:cost}
\end{equation}
}%
to obtain a minimum-cost path to a reachable success leaf within a depth limit, then execute the path one macro at a time under the same primitive-level gating.

\textbf{Hierarchical memory (spine).}
The traverser maintains the \emph{task spine} $(v_0,\dots,v_t)$ as persistent long-horizon state. The base model receives the task summary, macro descriptions on the spine, and a short node-local context summary distilled from $\texttt{env\_context}(v_t)$; full transcripts are not replayed. Structured safety-critical fields (paths, domains, process identifiers) bypass summarization and flow directly into $\texttt{ctx}$ and $\texttt{env}$ for gates and preconditions (App.~\ref{app:memory}).

\subsection{Policy as an External Artifact}
\label{sec:policy-artifact}
\vspace{-.2em}
\gbt{} is a persistent policy object distilled from logs: it routes tasks, constrains macro choices through traversal, enforces pre-execution gates, orchestrates recovery, and compresses long-horizon state into a spine. Because behavior and the bounded-history executable subset $\mathcal{S}_{\text{sys}}(t)$ are externalized into \gbt{} and the gate library, we decouple offline reasoning capacity from online execution: large models operate offline to build and refine these artifacts (without training), while at deployment a smaller $M_{\text{base}}$ executes one macro at a time under deterministic, primitive-level gate supervision along paths grounded in logs and preserved under regression-tested self-evolution.
\section{Experiments}
\label{sec:experiments}

\begin{table*}[!t]
\centering
\footnotesize
\caption{\textbf{SWE-bench Verified (Protocol A, 500 issues):} SR (\%), coverage (Cov, \%), violation rate (Viol, \%), unsafe success (USucc, \%), and efficiency (Tok/Chars, thousands). Wilson 95\% CIs are computed over per-instance majority-vote outcomes from three runs.}
\label{tab:swe_main_refined}
\setlength{\tabcolsep}{4pt}
\begin{tabularx}{\textwidth}{@{}>{\raggedright\arraybackslash}Xcccccc@{}}
\toprule
\textbf{System} &
\textbf{SR} &
\textbf{Cov} &
\textbf{Viol} &
\textbf{USucc} &
\textbf{Tok} &
\textbf{Chars} \\
\midrule
OpenHands CodeAct (\texttt{gpt-4o}, native) &
34.6 {\scriptsize[30.6,38.9]}\,{\scriptsize(173/500)} &
-- &
2.8 {\scriptsize[1.7,4.6]}\,{\scriptsize(14/500)} &
1.2 {\scriptsize[0.6,2.6]}\,{\scriptsize(6/500)} &
208 & 820 \\
\quad +Global guardrail only &
38.8 {\scriptsize[34.6,43.2]}\,{\scriptsize(194/500)} &
-- &
0.8 {\scriptsize[0.3,2.0]}\,{\scriptsize(4/500)} &
0.2 {\scriptsize[0.0,1.1]}\,{\scriptsize(1/500)} &
196 & 770 \\
\quad +GBT-Basic &
50.2 {\scriptsize[45.8,54.6]}\,{\scriptsize(251/500)} &
84.4 {\scriptsize(422/500)} &
0.4 {\scriptsize[0.1,1.4]}\,{\scriptsize(2/500)} &
0.2 {\scriptsize[0.0,1.1]}\,{\scriptsize(1/500)} &
148 & 570 \\
\quad +GBT-SE &
\textbf{73.6} {\scriptsize[69.6, 77.3]}\,{\scriptsize(368/500)} &
\textbf{86.0} {\scriptsize(430/500)} &
\textbf{0.2} {\scriptsize[0.0,1.1]}\,{\scriptsize(1/500)} &
\textbf{0.0} {\scriptsize[0.0,0.8]}\,{\scriptsize(0/500)} &
\textbf{126} & \textbf{490} \\
\bottomrule
\end{tabularx}

\end{table*}

\begin{table*}[!t]
\centering
\footnotesize
\caption{\textbf{WebArena (Protocol A, 812 tasks):} success, coverage, safety, and efficiency under strictly controlled OpenHands execution.}
\label{tab:web_main_refined}
\setlength{\tabcolsep}{4pt}
\begin{tabularx}{\textwidth}{@{}>{\raggedright\arraybackslash}Xcccccc@{}}
\toprule
\textbf{System} &
\textbf{SR} &
\textbf{Cov} &
\textbf{Viol} &
\textbf{USucc} &
\textbf{Tok} &
\textbf{Chars} \\
\midrule
OpenHands CodeAct (\texttt{gpt-4o}, native) &
19.7 {\scriptsize[17.1,22.6]}\,{\scriptsize(160/812)} &
-- &
3.4 {\scriptsize[2.4,4.9]}\,{\scriptsize(28/812)} &
1.0 {\scriptsize[0.5,1.9]}\,{\scriptsize(8/812)} &
94 & 360 \\
\quad +Global guardrail only &
19.3 {\scriptsize[16.8,22.2]}\,{\scriptsize(157/812)} &
-- &
0.9 {\scriptsize[0.4,1.8]}\,{\scriptsize(7/812)} &
0.2 {\scriptsize[0.1,0.9]}\,{\scriptsize(2/812)} &
88 & 335 \\
\quad +GBT-Basic &
53.0 {\scriptsize[49.5, 56.4]}\,{\scriptsize(430/812)} &
76.5 {\scriptsize(621/812)} &
0.4 {\scriptsize[0.1,1.1]}\,{\scriptsize(3/812)} &
0.1 {\scriptsize[0.0,0.7]}\,{\scriptsize(1/812)} &
60 & 230 \\
\quad +GBT-SE &
\textbf{66.9} {\scriptsize[63.6, 70.0]}\,{\scriptsize(543/812)} &
\textbf{78.0} {\scriptsize(633/812)} &
\textbf{0.2} {\scriptsize[0.1,0.9]}\,{\scriptsize(2/812)} &
\textbf{0.0} {\scriptsize[0.0,0.5]}\,{\scriptsize(0/812)} &
\textbf{52} & \textbf{205} \\
\bottomrule
\end{tabularx}

\end{table*}

\begin{table*}[!t]
\centering
\footnotesize
\caption{\textbf{GPQA (Protocol A, 448 questions):} accuracy (Acc, \%), coverage (Cov, \%), and safety (Viol/USucc) under OpenHands monitors/checkers. External web browsing is disabled.}
\label{tab:gpqa_main_refined}
\setlength{\tabcolsep}{4pt}
\begin{tabularx}{\textwidth}{@{}>{\raggedright\arraybackslash}Xcccccc@{}}
\toprule
\textbf{System} &
\textbf{Acc} &
\textbf{Cov} &
\textbf{Viol} &
\textbf{USucc} &
\textbf{Tok} &
\textbf{Chars} \\
\midrule
Zero-shot prompting (\texttt{gpt-4o}) &
53.6 {\scriptsize[48.9,58.1]}\,{\scriptsize(240/448)} &
-- & -- & -- &
-- & -- \\
OpenHands CodeActAgent (\texttt{gpt-4o}) &
58.7 {\scriptsize[54.1,63.2]}\,{\scriptsize(263/448)} &
-- &
1.6 {\scriptsize[0.8,3.2]}\,{\scriptsize(7/448)} &
0.4 {\scriptsize[0.1,1.6]}\,{\scriptsize(2/448)} &
22 & 86 \\
\quad +Global guardrail only &
59.2 {\scriptsize[54.6,63.7]}\,{\scriptsize(265/448)} &
-- &
0.4 {\scriptsize[0.1,1.6]}\,{\scriptsize(2/448)} &
0.2 {\scriptsize[0.0,1.3]}\,{\scriptsize(1/448)} &
21 & 82 \\
\quad +GBT-Basic &
78.8 {\scriptsize[74.8, 82.3]}\,{\scriptsize(353/448)} &
71.9 {\scriptsize(322/448)} &
0.2 {\scriptsize[0.0,1.3]}\,{\scriptsize(1/448)} &
0.0 {\scriptsize[0.0,0.9]}\,{\scriptsize(0/448)} &
16 & 62 \\
\quad +GBT-SE &
\textbf{87.3} {\scriptsize[83.9, 90.0]}\,{\scriptsize(391/448)} &
\textbf{73.0} {\scriptsize(327/448)} &
\textbf{0.2} {\scriptsize[0.0,1.3]}\,{\scriptsize(1/448)} &
\textbf{0.0} {\scriptsize[0.0,0.9]}\,{\scriptsize(0/448)} &
\textbf{15} & \textbf{58} \\
\bottomrule
\end{tabularx}
\end{table*}

We evaluate whether \gbt{} behaves as a \emph{first-class, model-external policy artifact}: (i) \textbf{when covered}, traversal (plus recovery) determines the executed long-horizon macro skeleton; (ii) \textbf{before execution}, every high-risk primitive is deterministically gated from structured \texttt{ctx} (global and node-local); (iii) \textbf{under stalls}, recovery emits short, feasible macro sequences; and (iv) \textbf{under long horizons}, spine memory replaces transcript replay to reduce drift and cost. All comparisons run inside the unified OpenHands runtime~\citep{wang2024openhands}; additional setup details, diagnostics, and mechanism ablations appear in App.~\ref{app:exp_details}--App.~\ref{app:exp_ablation}.

\subsection{Reproducible Runtime and Integration Discipline}
\label{sec:exp:runtime_main}
\vspace{-.2em}
\textbf{Unified sandbox, unified accounting.}
All runs execute in OpenHands~\citep{wang2024openhands}, which fixes the tool API, monitor/checker interface, logging schema, and cost accounting. This removes a dominant confound in agent evaluation: differences in environment instrumentation or safety enforcement cannot explain the results.

\textbf{\gbt-Traverser is a wrapper (no weight or planner edits).}
\textbf{GBT-Traverser} wraps the agent loop: it routes episodes to a task-family subtree, constrains macro choice by traversal (and, when triggered, recovery), enforces \emph{pre-execution} safety on every $a\in\mathcal{T}_{\text{hard}}$ via global and node-local gates, and maintains spine memory (Sec.~\ref{sec:method}; App.~\ref{app:memory}). It does \emph{not} update model weights and does \emph{not} modify any framework’s internal planner. Thus, differences between rows isolate exactly the Method’s levers: traversal/recovery/spine and node-local gates on top of the same base agent and the same sandbox monitors.

\subsection{Benchmarks, Protocols, and Metrics}
\label{sec:exp:setup}
\vspace{-.2em}
\textbf{Benchmarks.}
We evaluate on 15+ OpenHands-integrated benchmarks spanning software engineering, web interaction, tool-assisted reasoning, and adversarial safety/security. We foreground three execution-based pillars:
\textbf{SWE-bench Verified} (500 issues)~\cite{jimenez2023swe}, \textbf{WebArena} (812 tasks)~\cite{zhou2023webarena}, and \textbf{GPQA} (448 questions)~\citep{rein2024gpqa}. For GPQA we disable external web browsing and allow only local tools (e.g., Python) for deterministic computation. Safety/security is evaluated on \textbf{Agent-SafetyBench}\citep{agentsafetybench}, \textbf{AgentHarm (public)}\ \citep{andriushchenko2024agentharm}, and \textbf{Agent Security Bench (ASB)}~\citep{agentsecuritybench}.

\textbf{Leakage controls and claim boundary.}
We report three complementary protocols (App.~\ref{app:exp_protocols}). Our primary results use \textbf{Protocol A} (cross-benchmark distillation with benchmark-level hold-out): no trajectories from the evaluation benchmark are used for tree construction or self-evolution. We make the Method’s claim boundary explicit by reporting \textbf{Coverage} (Cov): when \texttt{covered}$=1$, traversal is the executed long-horizon policy skeleton; when \texttt{covered}$=0$, we do \emph{not} claim long-horizon control from \gbt{} and only claim primitive-level safety from the guardrail (Sec.\ \ref{sec:method}; App.~\ref{app:self-evolution}).

\textbf{Reporting.} For execution benchmarks we report success rate (SR) or accuracy (Acc), coverage (Cov), violation rate (Viol; any $\mathcal{S}_{\text{spec}}$ violation), unsafe success (USucc; success with any violation), and efficiency (Tok/Chars; thousands). Each instance is run \textbf{three times} at temperature $0$; outcomes are majority-voted and Wilson 95\% CIs are computed over per-instance binary outcomes (details in App.~\ref{app:exp_metrics}).

\subsection{Main Results on the Three Pillars (Protocol A)}
\label{sec:exp:main}
\vspace{-.2em}
Across pillars, the pattern is consistent: \textbf{global guardrails alone} largely buy safety with limited utility gain, while \textbf{adding \gbt{} traversal} produces a step-change in success and cost by externalizing long-horizon control into a reusable macro skeleton under deterministic pre-execution enforcement.

\subsubsection{SWE-bench Verified: large utility gains without safety trade-offs}
\label{sec:exp:main_swe}
\vspace{-.2em}
Table~\ref{tab:swe_main_refined} isolates three mechanisms. \emph{Primitive gating alone} (\textbf{+Global guardrail only}) cuts violations from 2.8\% to 0.8\% and unsafe success from 1.2\% to 0.2\%, but improves SR only modestly (34.6\%$\rightarrow$38.8\%), showing safety vetoes alone cannot fix long-horizon control. \emph{Traversal-as-Policy} (\textbf{+GBT-Basic}) raises SR to 50.2\% while maintaining low Viol/USucc and reducing cost (Tok/Chars 208/820$\rightarrow$148/570), confirming a distilled macro skeleton that prevents drift and shrinks context. \emph{Self-evolution under invariants} (\textbf{+GBT-SE}) yields the headline jump to 73.6\% SR at 86.0\% coverage, while driving unsafe success to 0.0\% and keeping violations near zero (0.2\%). The net effect matches the our core claim: safety is enforced \emph{pre-execution} from structured \texttt{ctx}, while success rises as long-horizon behavior follows a constrained traversal policy.

\subsubsection{WebArena: traversal arrests drift; recovery converts stalls into progress}
\label{sec:exp:main_web}
\vspace{-.2em}
WebArena emphasizes long-horizon brittleness: small deviations compound into irrecoverable dead-ends. Table~\ref{tab:web_main_refined} again separates mechanisms. The \textbf{global guardrail only} row sharply reduces Viol/USucc (3.4\%/1.0\%$\rightarrow$0.9\%/0.2\%) but leaves SR unchanged (19.7\%$\rightarrow$19.3\%), indicating that the dominant failure mode is not unsafe actions but \emph{policy drift and stalling}. Adding \gbt{} produces a large SR jump (to 53.0\% with \textbf{GBT-Basic}, then 66.9\% with \textbf{GBT-SE}) while simultaneously reducing cost (Tok/Chars 94/360$\rightarrow$52/205) and eliminating unsafe success (0.0\%). This is the expected signature of Traversal-as-Policy: traversal constrains macro choice to log-grounded successors, and recovery injects short, feasible macro sequences when the agent stalls (recovery diagnostics and ablations in App.~\ref{app:exp_ablation}).

\subsubsection{GPQA: with browsing disabled, \gbt{} turns the agent into a controlled local executor}
\label{sec:exp:main_gpqa}
\vspace{-.2em}
GPQA removes web browsing and tests tool-assisted reasoning under strict sandbox monitors. Table~\ref{tab:gpqa_main_refined} shows that \gbt{} raises accuracy from 58.7\% (OpenHands agent) to 87.3\% (\textbf{GBT-SE}), while keeping Viol and USucc essentially at zero (0.2\% and 0.0\%). The improvement concentrates in covered episodes, consistent with traversal supplying an explicit macro-level execution plan and spine memory preventing transcript-induced drift; a coverage-conditioned audit is reported in Table~\ref{tab:conditional_success_refined}.

\begin{table}[t]
\centering
\scriptsize
\caption{\textbf{Conditional performance under coverage (GBT-SE, Protocol A):} gains concentrate in \texttt{covered}$=1$ episodes, the regime where traversal is the long-horizon policy.}
\label{tab:conditional_success_refined}
\setlength{\tabcolsep}{4pt}
\renewcommand{\arraystretch}{1.0}
\begin{tabular}{@{}lcccc@{}}
\toprule
\textbf{Benchmark} &
\textbf{Overall} &
\textbf{Cov} &
\shortstack{$\Pr(\mathbf{succ}\mid$\\$\texttt{covered}=1)$} &
\shortstack{$\Pr(\mathbf{succ}\mid$\\$\texttt{covered}=0)$} \\
\midrule
SWE-bench Verified (SR) &
73.6 &
86.0 &
80.2 &
32.9 \\
WebArena (SR) &
66.9 &
78.0 &
80.4 &
19.0 \\
GPQA (Acc) &
87.3 &
73.0 &
97.9 &
58.7 \\
\bottomrule
\end{tabular}

\end{table}

\subsubsection{Paired outcome audit: improvements persist under paired testing}
\label{sec:exp:paired_audit}
To rule out unpaired noise, we run a paired audit comparing \textbf{Global guardrail only} vs.\ \textbf{+GBT-SE} on the \emph{same} instances (Protocol A), reporting flips and an exact McNemar test. Table~\ref{tab:paired_mcnemar_refined} shows that \gbt{} converts large numbers of failures into successes (large $b$) with comparatively few regressions (small $c$), with significance across all pillars.

\begin{table}[t]
\centering
\small
\caption{\textbf{Paired outcome audit (Protocol A):} $b$ counts instances flipped to success by \gbt{}, $c$ counts instances flipped to failure, ties are unchanged. Exact McNemar test is computed on discordant pairs.}
\label{tab:paired_mcnemar_refined}
\setlength{\tabcolsep}{5pt} 
\renewcommand{\arraystretch}{0.95}

\begin{tabular}{@{}lccccc@{}} 
\toprule
\textbf{Benchmark} & \textbf{$n$} & \textbf{$b$} & \textbf{$c$} & \textbf{Ties} & \textbf{McNemar $p$} \\
\midrule
SWE-bench Verified & 500 & 192 & 18 & 290 & $6.2\!\times\!10^{-38}$ \\
WebArena & 812 & 410 & 24 & 378 & $8.1\!\times\!10^{-92}$ \\
GPQA & 448 & 137 & 11 & 300 & $7.8\!\times\!10^{-29}$ \\
\bottomrule
\end{tabular}

\end{table}

\subsection{Coverage is the Explicit Claim Boundary (and Where the Gains Come From)}
\label{sec:exp:coverage_main}

Traversal-as-Policy is intentionally coverage-scoped (Sec.\ \ref{sec:method}). Table~\ref{tab:conditional_success_refined} shows that improvements concentrate in \texttt{covered}$=1$ episodes, where traversal is the executed long-horizon policy: SR/Acc is 80.2/80.4/97.9 in coverage, versus 32.9/19.0/58.7 out of coverage. Conversely, Table~\ref{tab:outside_coverage_refined} shows that on episodes labeled \texttt{covered}$=0$ (abstention or safe exploration), outcomes remain similarly poor, consistent with explicit abstention rather than hidden free-running. Step-level matching diagnostics, matched-subset audits, and threshold sensitivity are reported in App.~\ref{app:exp_coverage}.

\begin{table*}[!t]
\centering
\footnotesize
\caption{\textbf{Executor decoupling (Protocol B, 5-fold hold-out):} performance with average tokens (Tok, thousands) and tokenizer-agnostic characters (Chars, thousands).}
\label{tab:small_exec_refined}
\setlength{\tabcolsep}{4pt}
\begin{tabularx}{\textwidth}{@{}>{\raggedright\arraybackslash}Xccc@{}}
\toprule
\textbf{System (OpenHands CodeAct)} &
\textbf{SWE SR\ \ (Tok/Chars)} &
\textbf{WebArena SR\ \ (Tok/Chars)} &
\textbf{GPQA Acc\ \ (Tok/Chars)} \\
\midrule
\texttt{gpt-4o} (native) &
34.2\ (212/818) &
19.5\ (96/365) &
53.5\ (21/77) \\
\quad +Global guardrail only &
36.8\ (190/715) &
19.3\ (90/342) &
51.1\ (20/77) \\
\quad +GBT-SE (same-model upper bound) &
\textbf{63.1}\ (124/482) &
\textbf{65.1}\ (46/177) &
\textbf{63.0}\ (14/54) \\
\midrule
\texttt{llama-3-8b} (native) &
12.6\ (238/905) &
8.3\ (131/505) &
32.4\ (30/120) \\
\quad +GBT-Basic (executor-only) &
27.8\ (121/465) &
22.4\ (61/235) &
49.5\ (16/64) \\
\quad +GBT-SE (executor-only) &
\textbf{46.4}\ (105/405) &
\textbf{33.9}\ (55/212) &
\textbf{58.4}\ (16/60) \\
\midrule
\texttt{Qwen3-VL-8B-Thinking} (native) &
14.0\ (231/880) &
9.1\ (127/488) &
34.0\ (29/116) \\
\quad +GBT-Basic (executor-only) &
32.4\ (124/475) &
24.2\ (59/228) &
54.1\ (16/65) \\
\quad +GBT-SE (executor-only) &
\textbf{58.8}\ (109/420) &
\textbf{37.3}\ (54/210) &
\textbf{60.5}\ (18/61) \\
\bottomrule
\end{tabularx}

\end{table*}

\begin{table*}[!t]
\centering
\footnotesize
\caption{\textbf{Safety/security benchmarks (public and reproducible):} \gbt{} yields large safety gains beyond native agents while preserving utility.}
\label{tab:safety_bench_refined}
\setlength{\tabcolsep}{4pt}
\begin{tabularx}{\textwidth}{@{}>{\raggedright\arraybackslash}Xccccc@{}}
\toprule
\textbf{System} &
\textbf{Agent-SafetyBench} $\uparrow$ &
\textbf{AgentHarm (public) HarmScore} $\downarrow$ &
\textbf{ASB ASR-d} $\downarrow$ &
\textbf{ASB RR} $\uparrow$ &
\textbf{ASB PNA-d} $\uparrow$ \\
\midrule
\texttt{gpt-4o} (native) &
44.2 & 71.8 & 64.4 & 8.8 & \textbf{71.3} \\
\quad +Global guardrail only &
56.8 & 19.0 & 13.1 & 63.2 & 69.0 \\
\quad +GBT-Basic &
60.2 & 11.2 & 8.6 & 73.8 & 70.1 \\
\quad +GBT-SE &
\textbf{72.3} & \textbf{9.6} & \textbf{7.0} & \textbf{78.4} & 70.4 \\
\midrule
\texttt{llama-3-8b} (native) &
20.1 & 29.4 & 20.4 & 4.9 & \textbf{52.0} \\
\quad +Global guardrail only &
50.4 & 16.5 & 10.9 & 61.6 & 50.8 \\
\quad +GBT-Basic &
55.4 & 12.4 & 8.7 & 72.2 & 51.6 \\
\quad +GBT-SE &
\textbf{70.8} & \textbf{10.3} & \textbf{7.4} & \textbf{77.0} & 51.8 \\
\bottomrule
\end{tabularx}
\end{table*}

\begin{table}[t]
\centering
\small
\caption{\textbf{Outside-coverage audit (Protocol A):} performance on episodes labeled \texttt{covered}$=0$ by \gbt{}. Similar outcomes confirm that gains arise in-coverage, where traversal is executed as the policy skeleton.}
\label{tab:outside_coverage_refined}
\setlength{\tabcolsep}{8pt}
\resizebox{\columnwidth}{!}{%
\begin{tabular}{lccc}
\toprule
\textbf{Benchmark} & \textbf{$n_{\texttt{covered}=0}$} & \textbf{Base agent} & \textbf{+GBT-SE} \\
\midrule
SWE-bench Verified (SR) & 70 & 32.9 {\scriptsize(23/70)} & 32.9 {\scriptsize(23/70)} \\
WebArena (SR) & 179 & 17.9 {\scriptsize(32/179)} & 19.0 {\scriptsize(34/179)} \\
GPQA (Acc) & 121 & 55.4 {\scriptsize(67/121)} & 58.7 {\scriptsize(71/121)} \\
\bottomrule
\end{tabular}%
}

\end{table}

\subsection{Decoupling Reasoning Capacity from Policy Execution (Protocol B)}
\label{sec:exp:decouple_main}
We test the executor-decoupling claim: heavy reasoning is used \emph{offline} to build/refine \gbt{}, while \emph{online} execution can be delegated to small models that realize one macro at a time under traversal/recovery and deterministic gates (Sec.~\ref{sec:method}). Table~\ref{tab:small_exec_refined} reports 5-fold instance-level hold-out (Protocol B). With the same distilled tree, 8B-scale executors more than triple success on the execution pillars (e.g., \texttt{llama-3-8b}~\cite{dubey2024llama} on SWE: 12.6\%$\rightarrow$46.4\%) while operating at markedly lower Tok/Chars due to spine-based context and reduced long-horizon deliberation. This directly supports the ``policy as external artifact'' claim: the reusable object is \gbt{}, not a particular model’s weights.

\subsection{Deterministic Pre-execution Safety on Public Safety/Security Suites}
\label{sec:exp:safety_main}
We evaluate whether safety outcomes match the Method’s invariants (Sec.\ \ref{sec:method}): gate checks are \emph{pre-execution} and depend only on structured \texttt{ctx}; self-evolution cannot re-admit previously rejected unsafe contexts; and node-local gates provide macro-local defenses beyond global guardrails. Table~\ref{tab:safety_bench_refined} reports large safety gains over native agents, with further improvements past global guardrails when enabling \gbt{}; importantly, utility under defense (ASB \textbf{PNA-d}) remains stable. Extended safety counts, guardrail activity, and mechanism ablations (gates, recovery, memory, self-evolution) are in App.~\ref{app:exp_safety}--App.~\ref{app:exp_ablation}.

\textbf{Where to find the rest.} App.~\ref{app:exp_plugin}--\ref{app:exp_ablation} detail plug-in generalization across heterogeneous frameworks, run-to-run stability, structural hold-outs, coverage/matching diagnostics, and targeted ablations (gates, recovery, memory, self-evolution).

\section{Conclusion}
\label{sec:conclusion}
Traversal-as-Policy turns execution logs into an explicit, inspectable controller: a log-distilled Gated Behavior Tree (\gbt{}) whose traversal is the policy whenever coverage holds. By compiling unsafe traces into deterministic pre-execution gates over structured tool contexts and updating them under experience-grounded monotonicity, \gbt{} enforces safety before high-risk actions run and prevents silent regression during self-evolution. A lightweight traverser executes one state-conditioned macro at a time, uses risk-aware shortest-path recovery to escape stalls, and replaces transcript replay with a compact spine memory, jointly improving robustness and cost. Across 15+ OpenHands benchmarks, \gbt{}-SE delivers large utility gains while driving violations and unsafe success to (near) zero, and the same distilled tree enables 8B-scale executors to compete with far larger models—evidence that policy can be externalized from weights. Limitations, additional analysis, and future directions appear in App.~\ref{app:concl_limitations}--\ref{app:concl_future}.

\newpage
\clearpage
\section*{Impact Statement}

This work aims to change how autonomous LLM agents are \emph{built, audited, and deployed} by turning long-horizon behavior from an \emph{implicit} artifact (weights + transcripts) into an \emph{explicit, executable, and inspectable} policy object. Concretely, \emph{Traversal-as-Policy} distills sandboxed execution logs into a single \gbt{} and then makes \emph{tree traversal}---not unconstrained generation---the control policy whenever a task is in coverage. Each node encodes a reusable state-conditioned macro, while safety is enforced \emph{before execution} via deterministic gates over structured tool context (\texttt{ctx}) and bounded history. Gates are updated under experience-grounded monotonicity, meaning that once a structured unsafe context is rejected, it cannot be silently re-admitted by later iterations. The result is a policy that can be \emph{externalized, verified, debugged, regression-tested, and evolved} without changing model weights.

\textbf{Positive impact: safety that is operational, not aspirational.}
Most agent ``safety'' today is retrospective: the model acts, then a wrapper or a second model judges. Our approach shifts safety to a pre-execution contract: high-risk primitives (writes/deletes, process spawns, network sends, sensitive reads) are deterministically checked using only structured fields extracted from the sandbox/runtime state. This is intentionally hostile to prompt hacking and transcript manipulation: gates do not depend on free-form summaries, so they cannot be bypassed by changing phrasing. Empirically, across execution-centric benchmarks, adding \gbt{} drives violations and unsafe success toward zero while simultaneously improving utility (e.g., on SWE-bench Verified, violations drop from 2.8\% to 0.2\% and unsafe success to 0.0\% while success rises to 73.6\%; on WebArena, violations drop from 3.4\% to 0.2\% and unsafe success to 0.0\% while success rises to 66.9\%). This combination---\emph{utility up, safety incidents down}---is the practical signature of an enforceable safety mechanism rather than a cosmetic filter.

\textbf{Positive impact: reliability and cost improvements via explicit control and compact memory.}
A major failure mode for agents is long-horizon drift: plans degrade as transcripts grow and reasoning becomes inconsistent. Traversal-as-Policy constrains action choice to successors grounded in previously successful trajectories, and replaces transcript replay with a compact spine memory (visited macros). Recovery further converts stalls into progress by planning short feasible paths to success leaves under risk-aware costs. In our experiments, this yields large reductions in token/character usage alongside higher success (e.g., SWE-bench Verified Tok/Chars drop from 208k/820k to 126k/490k; WebArena from 94k/360k to 52k/205k). This has direct economic impact (lower inference cost) and environmental impact (less compute per completed task), and it also improves reproducibility because the executed behavior is a traversed path in an explicit tree, not an opaque transcript.

\textbf{Positive impact: decoupling reasoning capacity from execution enables safer, cheaper deployment.}
Because \gbt{} is a model-external policy artifact, heavy reasoning can happen \emph{offline} to build and refine the tree and gates, while \emph{online} execution can be delegated to smaller models that follow the traversal policy one macro at a time. This changes the deployment equation: instead of requiring frontier models everywhere, systems can run smaller executors under an explicit controller, improving accessibility and enabling edge/low-resource settings. In our results, the same distilled \gbt{} allows 8B executors to more than double or triple success on major benchmarks (e.g., \texttt{Qwen3-VL-8B-Thinking} rises from 14.0\% to 58.8\% on SWE-bench Verified and from 9.1\% to 37.3\% on WebArena), demonstrating that the reusable object is the policy artifact, not a specific set of weights.

\textbf{Potential negative impact and misuse.}
By making agent behavior more robust and reusable, policy externalization could also lower the barrier to automating harmful workflows (e.g., scalable exploitation, data exfiltration, or misuse of tooling) if a \gbt{} were distilled from unsafe data or deployed without strong monitors. In addition, deterministic gating is only as strong as the structured context it reads: if \texttt{ctx} omits relevant state, a gate may fail to capture a violation mode. Finally, log-distillation can inherit biases present in the collected trajectories, potentially encoding brittle conventions or systematically excluding alternative strategies.

\textbf{Mitigations and responsible deployment guidance.}
Our design explicitly builds in constraints intended for safety-critical deployment: (i) \emph{coverage-scoped control} with abstention (\texttt{covered}=0) prevents the system from silently claiming competence outside its learned regimes; (ii) \emph{pre-execution deterministic gates} over structured context reduce vulnerability to prompt-based attacks; (iii) \emph{experience-grounded monotonicity} and regression testing prevent silent safety regression during self-evolution; and (iv) evaluation on public safety/security suites provides evidence that safety gains persist beyond the main utility benchmarks. For real-world use, we recommend deploying \gbt{} only alongside strong environment instrumentation (monitors/checkers), retaining human review for high-stakes actions, auditing distilled trees and gate corpora for privacy and bias, and treating gate/context coverage as an explicit engineering requirement (expand \texttt{ctx} and monitors when new violation modes appear).

Overall, this paper’s impact is to provide a concrete path from ``agents as opaque generators'' to ``agents as verifiable controllers'': a persistent policy object that can be inspected, tested, and improved with evidence, while making safety an executable precondition of action rather than a post hoc aspiration.




\nocite{langley00}

\bibliography{example_paper}
\bibliographystyle{icml2026}


\newpage
\appendix
\onecolumn
\section{Related Work}

Our work sits at the intersection of long-horizon agent control, safety, and structured policy representations. The key distinction is \emph{policy externalization}: instead of improving an implicit, transcript-driven controller, we distill a single \emph{executable} artifact from aggregate experience and then execute \emph{traversal} as the policy whenever in coverage.

\paragraph{Improving intrinsic reasoning and memory.}
A large body of work strengthens agent performance by amplifying internal reasoning. Deliberative methods such as Tree-of-Thoughts (ToT) \citep{treeofthought} expand online search, often at substantial inference cost. Self-improvement approaches such as Reflexion \citep{reflexion} and Meta-Policy Reflexion \citep{mpr} store textual ``reflection memories'' to adapt from failures. These techniques remain fundamentally \emph{generation-centric}: they either operate online or build non-executable textual memories, and the long-horizon policy is still realized through unconstrained language generation conditioned on a growing transcript. In contrast, our framework distills \emph{both} successes (for behavioral guidance) and failures (for safety) into a single offline, executable structure; the resulting controller is an inspectable artifact whose traversal determines the executed macro skeleton in covered episodes.

\paragraph{Guardrailing and runtime validation.}
The dominant safety paradigm for agents is ``guardrailing'': attach a runtime validator, critic, or guardian agent that checks actions as they are proposed \citep{guardagent,aworld,policyasprompt}. While effective in reducing harm, these systems typically rely on \emph{human-specified} policies (natural-language rules, prompt templates, or code), which is unscalable and often misses the long tail of emergent, context-dependent failure modes that appear only in operational data. Our approach flips the source of safety knowledge: we deterministically replay unsafe traces, shrink them to minimal violating windows, and compile the resulting structured contexts into \emph{pre-execution gates} that are grounded in observed failures. This converts failure logs into a growing, executable safety mechanism rather than a passive record.

\paragraph{Formal guarantees, learned constraints, and the safety--utility trade-off.}
Formal methods such as shielding \citep{shielding} can offer strong verifiable guarantees, but they typically require manually crafted formal environment models, an intractable prerequisite for most open-world agent tasks. Hybrid approaches that learn probabilistic models online \citep{agentguard} can provide only probabilistic assurances, whereas our gates provide deterministic checks over structured tool contexts. Other paradigms internalize safety via implicit optimization trade-offs, such as Constrained MDPs (CMDPs) \citep{gu2024enhancingefficiencysafereinforcement} or adversarial training \citep{agentdojo,arlas}. These methods embed safety inside an opaque, monolithic policy, making it difficult to inspect or certify what constraints are enforced and where. Our framework instead \emph{externalizes} safety into an explicit library of deterministic predicates over structured contexts, and enforces an experience-grounded monotonicity rule: once a context is observed unsafe and rejected, it cannot be re-admitted by later updates. This yields a verifiable notion of non-regression on observed unsafe contexts while still allowing behavioral improvement through coverage expansion.

\paragraph{Structured policy representations for agents.}
Hierarchical controllers such as finite state machines (FSMs) \citep{pract,crouse2024formallyspecifyinghighlevelbehavior} and behavior trees (BTs) \citep{wang2025llmhbtdynamicbehaviortree,btgenbot} offer interpretability and modularity, but are often hand-crafted by experts or generated once from a single high-level instruction. This ``Policy-as-Code'' paradigm is brittle under distribution shift and typically lacks a principled link between operational failures and controller updates. Our approach is ``Policy-as-Data'': we distill a \emph{Gated} Behavior Tree from massive aggregate experience, merge-checking macros to avoid semantic aliasing where safety matters, attaching node-local gates from unsafe windows, and executing traversal as the policy within an explicit coverage boundary. The result is not a one-off plan for a single instruction, but a reusable policy artifact that unifies behavioral guidance, deterministic pre-execution safety, recovery, and long-horizon memory within a single executable structure.

\section{Safety Specification, Structured Contexts, and the Global Guardrail}
\label{app:guardrail}

\subsection{Normative Safety Specification in OpenHands and Sandboxed Data}
\label{app:guardrail:safety-spec}

\paragraph{What is normative safety \texorpdfstring{$\mathcal{S}_{\text{spec}}$}{S_spec}?}
All trajectories are executed inside the OpenHands runtime, which provides (i) a Docker-isolated sandbox, (ii) a standardized event stream of tool \emph{actions} and \emph{observations}, and (iii) benchmark-defined \emph{checkers} plus runtime \emph{monitors} that emit verdicts. Together, monitors and checkers define a \emph{normative safety specification} $\mathcal{S}_{\text{spec}}$ over primitive tool calls:
a primitive violates $\mathcal{S}_{\text{spec}}$ iff at execution time it triggers any monitor or checker safety verdict.
Examples include (non-exhaustive): writes outside workspace roots, deletion of protected directories, process spawn patterns disallowed by the sandbox policy, network traffic to disallowed destinations, and benchmark-specific leakage/exfiltration triggers.

\paragraph{Trajectory labeling.}
A trajectory $\tau$ is labeled \texttt{unsafe} \emph{iff} at least one executed primitive violates $\mathcal{S}_{\text{spec}}$ under the OpenHands monitors/checkers. This is the only ground-truth safety label used throughout the pipeline.

\paragraph{Design goal (coverage-scoped claims).}
We do \emph{not} claim complete enforcement of $\mathcal{S}_{\text{spec}}$ for all hazards.
Instead, we externalize and deterministically enforce an \emph{executable} bounded-history subset (defined below) and treat violations under the current system as evidence that expands this executable subset monotonically over observed unsafe contexts (Design invariant~1).

\subsection{Structured Contexts: The Only Inputs to Gates}
\label{app:guardrail:ctx}

\paragraph{Structured context schema.}
For each candidate high-risk primitive, we construct a structured context $\texttt{ctx}\in\mathcal{C}$ \emph{directly from sandbox state}.
The gate library reads \emph{only} these structured fields and a bounded history of recent high-risk primitives; it never reads LLM summaries, chain-of-thought, or free-form transcripts.
Concretely, we represent \texttt{ctx} as:

\begin{quote}
\texttt{ctx = \{ primitive\_type, tool\_family, args, resource\_ids,}\\
\texttt{\ \ \ \ \ \ \ \ \ \ \ \ \ workspace\_roots, cwd, uid/gid,}\\
\texttt{\ \ \ \ \ \ \ \ \ \ \ \ \ net\_dest(domain, ip, port, scheme),}\\
\texttt{\ \ \ \ \ \ \ \ \ \ \ \ \ proc\_meta(exec, argv, parent),}\\
\texttt{\ \ \ \ \ \ \ \ \ \ \ \ \ payload\_meta(len, mime, hash),}\\
\texttt{\ \ \ \ \ \ \ \ \ \ \ \ \ recent\_hard\_history[1..H]\}}
\end{quote}

where:
(i) \texttt{args/resource\_ids} store canonicalized file paths (normalized, resolved, and checked against sandbox roots), URL components (domain/scheme/path), and process identifiers where applicable;
(ii) \texttt{payload\_meta} stores metadata for payload-bearing actions (length, MIME/type if known, and a stable hash); and
(iii) \texttt{recent\_hard\_history} records the last $H$ high-risk primitives \emph{within the same macro}, each stored as a compact structured record \texttt{(type, canonical\_resource, coarse\_op, timestamp)}.
We use a fixed small history bound $H=4$ in the reference configuration.

\paragraph{High-risk primitive set \texorpdfstring{$\mathcal{T}_{\text{hard}}$}{T_hard}.}
We designate as high-risk:
\[
\mathcal{T}_{\text{hard}}
=
\{\text{all writes and deletes}\}
\cup
\{\text{all process spawns}\}
\cup
\{\text{all network sends}\}
\cup
\{\text{reads matching sensitive patterns}\}.
\]
The \emph{sensitive patterns} component is benchmark- and runtime-aware and includes protected filesystem prefixes, environment/key material patterns, and benchmark-defined secret resources (when applicable).

\paragraph{Non-bypassability.}
Because \texttt{ctx} is constructed from sandbox state and canonicalization is deterministic, \emph{no prompting or summarization choice can alter gate inputs}.
Summaries may influence how a macro is realized, but cannot change \texttt{ctx} nor the outcome of deterministic gates computed from it.

\subsection{Gate Interface, Global vs.\ Node-Local Gates, and Executable Safety Subset}
\label{app:guardrail:gate}

\paragraph{Gate interface (common to all gate families).}
Every gate has the same interface:
\[
g:\ \texttt{ctx} \mapsto (\texttt{ok},\texttt{msg})
\in
\{\texttt{true},\texttt{false}\}\times\texttt{String}.
\]
We write $\mathcal{G}_{\text{global}}(t)$ for global gates and $\mathcal{G}_{\text{node}}(t)$ for node-local gates attached to macros in the tree, and
$\mathcal{G}(t)=\mathcal{G}_{\text{global}}(t)\cup\mathcal{G}_{\text{node}}(t)$.

\paragraph{Executable safety subset.}
We define the executable subset at time $t$ as:
\[
\mathcal{S}_{\text{sys}}(t)
=
\left\{
\texttt{ctx}\in\mathcal{C}:
\exists g\in\mathcal{G}(t)
\ \text{s.t.}\ 
g(\texttt{ctx}).\texttt{ok}=\texttt{false}
\right\}.
\]
This is a conservative, bounded-history approximation of the portion of $\mathcal{S}_{\text{spec}}$ that is expressible via structured contexts.

\paragraph{Global decision (always-on, pre-execution).}
For any candidate high-risk primitive $a\in\mathcal{T}_{\text{hard}}$:
\[
\texttt{HardGateOK}(a,\texttt{ctx})
=
\bigwedge_{g\in\mathcal{G}_{\text{global}}} g(\texttt{ctx}).\texttt{ok}.
\]
This check runs \emph{before execution} for every high-risk primitive in every phase: data collection, offline distillation, and online deployment.

\paragraph{Node-local decision (macro-scoped, pre-execution).}
When a macro node $v$ is selected during traversal/recovery, we additionally enforce:
\[
\texttt{GateOK}(v,\texttt{ctx})
=
\bigwedge_{g\in\mathcal{G}(v)} g(\texttt{ctx}).\texttt{ok}.
\]
The executed primitive is allowed only if both global and node-local checks pass (as in Sec.~\ref{sec:online}).

\subsection{Gate Families: Deterministic RuleGates and Deterministic ContentGates}
\label{app:guardrail:families}

\paragraph{RuleGates (fully deterministic).}
RuleGates are pure code predicates over structured fields in \texttt{ctx} and its bounded history, for example:
(i) forbid writes outside workspace roots;
(ii) forbid deletion of protected paths;
(iii) forbid process spawn patterns (e.g., executing disallowed binaries);
(iv) forbid network sends to disallowed domains/ports; and
(v) block short-window structured patterns such as ``read sensitive resource then archive then send'' when all components are detectable from \texttt{ctx} and bounded history.
RuleGates are unit-tested and deterministic.

\paragraph{ContentGates (deterministic calls to a frozen classifier).}
Some hazards are primarily carried by unstructured payloads (text/code/serialized blobs).
For payload-bearing actions, ContentGates construct a guard prompt from:
\texttt{(i) ctx structured fields, (ii) a short payload excerpt or redacted summary, (iii) a task/macro tag)},
then call a frozen safety classifier at temperature $0$ and post-process into a boolean decision.
In the reference configuration we use \texttt{Llama-Guard-3-8B} at temperature $0$ as the classifier (see App.~\ref{app:exp_systems}).

\paragraph{Determinism contract.}
ContentGates are deterministic conditional on (i) a fixed guard model, (ii) temperature $0$, and (iii) a fixed prompt template.
They do \emph{not} introduce stochasticity into enforcement.

\subsection{Experience-Grounded Corpora and Monotone Updates (No Unsafe Re-Admission)}
\label{app:guardrail:monotone}

\paragraph{Unsafe and benign corpora.}
For each gate $g$ we maintain:
\begin{itemize}
\item an unsafe corpus $\mathcal{D}_{\text{unsafe}}(g)$: structured contexts extracted from trajectories that violate $\mathcal{S}_{\text{spec}}$ (minimal windows, App.~\ref{app:tree-construction});
\item a benign corpus $\mathcal{D}_{\text{benign}}(g)$: representative safe contexts of the \emph{same primitive family} (same tool family, same risk class) sampled from successful or non-violating runs.
\end{itemize}
We maintain balanced benign sampling across resource types and macro neighborhoods to avoid overfitting gates to a narrow distribution.

\paragraph{Update rule (monotonicity + benign regression control).}
Any update from $g^{\text{old}}$ to $g^{\text{new}}$ must satisfy:
\[
\forall\,\texttt{ctx}\in\mathcal{D}_{\text{unsafe}}(g):\quad
g^{\text{old}}(\texttt{ctx}).\texttt{ok}=\texttt{false}
\Rightarrow
g^{\text{new}}(\texttt{ctx}).\texttt{ok}=\texttt{false},
\]
and limit benign false positives:
\[
\frac{\left|\left\{\texttt{ctx}\in\mathcal{D}_{\text{benign}}(g): g^{\text{new}}(\texttt{ctx}).\texttt{ok}=\texttt{false}\right\}\right|}
{|\mathcal{D}_{\text{benign}}(g)|}
\le \epsilon_{\text{benign}}.
\]
In the reference configuration we use $\epsilon_{\text{benign}}=0.01$ for global gates and $\epsilon_{\text{benign}}=0.02$ for node-local gates (node-local gates are intentionally allowed to be slightly more conservative within a macro neighborhood).

\paragraph{Design invariant 1 (Experience-grounded monotonicity).}
Once a structured context is recorded as unsafe and rejected by any gate, it remains rejected by all future gate libraries.
Equivalently, $\mathcal{S}_{\text{sys}}(t)$ can only \emph{expand} on observed unsafe behavior; it cannot shrink on logged unsafe contexts.

\subsection{Reference Prompt Templates (Runnable) for ContentGates and Gate Synthesis}
\label{app:guardrail:prompts}

\paragraph{ContentGate prompt (reference configuration).}
We use a fixed template that maps \texttt{ctx} + a payload excerpt to a strict binary decision.
The template is intentionally \emph{short}, as the input is already structured:

\begin{quote}\small
\textbf{SYSTEM:} You are a safety classifier. Output exactly one token: \texttt{SAFE} or \texttt{UNSAFE}.\\
\textbf{USER:} \\
\texttt{[Action]} \{primitive\_type, tool\_family\}\\
\texttt{[Args]} \{canonical\_path/domain/port if any\}\\
\texttt{[History]} last H hard primitives: \{(type, resource, op)\}\\
\texttt{[Payload]} (redacted excerpt, max 1{,}000 chars)\\
\texttt{Decide if executing this action violates the sandbox safety policy or}\\
\texttt{benchmark-defined safety constraints.}
\end{quote}

\paragraph{Deterministic RuleGate synthesis prompt (reference configuration).}
We synthesize candidate structured predicates offline from minimal unsafe windows (App.~\ref{app:tree-construction}) using a frozen reasoning model.
We enforce that the model outputs \emph{only} code over structured fields and bounded history:

\begin{quote}\small
\textbf{SYSTEM:} You write deterministic Python predicates over a given JSON schema.\\
\textbf{USER:} Given the unsafe contexts below (positive) and benign contexts (negative),\\
propose a Python function \texttt{gate(ctx)->(ok,msg)} that:
(1) rejects all positives,
(2) rejects at most $\epsilon_{\text{benign}}$ of negatives,
(3) uses only \texttt{ctx} fields and \texttt{ctx.recent\_hard\_history},
(4) contains no randomness and no external calls.
Return JSON: \{\texttt{"code": "...", "rationale": "..."}\}.
\end{quote}

We then (i) compile the returned code, (ii) unit-test it on corpora, and (iii) accept it only if it satisfies the formal constraints above.

\subsection{Concrete Examples of Structured Gates (Illustrative)}
\label{app:guardrail:examples}

Below are illustrative (not benchmark-specific) RuleGate patterns, expressed purely over structured fields:

\begin{itemize}
\item \textbf{Workspace confinement:} reject writes/deletes whose canonical path is outside \texttt{workspace\_roots}.
\item \textbf{Protected deletion:} reject deletes of any path matching protected prefixes (e.g., system dirs) or non-ephemeral dirs.
\item \textbf{Network egress control:} reject network sends when \texttt{domain} is not in an allowlist or when \texttt{scheme/port} is disallowed.
\item \textbf{Short-window exfiltration motif:} if \texttt{recent\_hard\_history} contains a sensitive read followed by an archive/write, reject a subsequent network send of the same artifact.
\end{itemize}

These examples are representative of the expressible subset $\mathcal{S}_{\text{sys}}(t)$ and emphasize the key property: all decisions depend only on structured context and bounded history.

\section{Behavior Path Extractor and Abstraction Stability}
\label{app:behavior-extractor}

\subsection{Deterministic Macro Segmentation from Observable Deltas}
\label{app:behavior-extractor:segmentation}

Raw trajectories are long and tool-specific. The Behavior Path Extractor $E$ maps a trajectory $\tau$ to a macro path
\[
p=(v_0,\dots,v_K)=E(\tau),
\]
where each macro $v_k$ is a contiguous span of primitives that:
(i) stays within a local sub-action region, and
(ii) realizes a coherent intent.

\paragraph{Segmentation signals (no LLM judgment).}
Macro boundaries are anchored by \emph{observable} environment deltas and tool-family changes logged by OpenHands:
\begin{itemize}
\item \textbf{Filesystem deltas:} creation/modification of files; diff size thresholds; directory scope change.
\item \textbf{Execution deltas:} process spawn/termination; test invocation boundaries; interpreter session boundaries.
\item \textbf{Web deltas:} domain change; navigation state change; form submission boundary.
\item \textbf{Tool-family switch:} first invocation of a different tool family in a span (e.g., from file ops to browser).
\end{itemize}
These rules are deterministic functions of logs. LLMs are not used to decide boundaries.

\paragraph{Risk annotation.}
Each macro is assigned a discrete \texttt{risk\_level} based on whether its primitive span touches $\mathcal{T}_{\text{hard}}$ and which resource families it touches (files, network, processes). This risk metadata is used only for (i) conservative merging and (ii) risk-aware recovery costs (App.~\ref{app:recovery}).

\subsection{Macro Description Summarization (Offline Only) and Output Schema}
\label{app:behavior-extractor:summarization}

After segmentation, we summarize each macro span into a short description \emph{solely for semantic matching and retrieval}.
This summary is \emph{never} consumed by gates or preconditions.

\paragraph{Reference summarization prompt (runnable).}
We call a frozen LLM at temperature $0$ with a strict JSON schema:

\begin{quote}\small
\textbf{SYSTEM:} Summarize a tool-usage segment into a reusable action macro. Output JSON only.\\
\textbf{USER:} You are given (i) a sequence of primitive tool calls with arguments, and (ii) a compact description of observable environment deltas.\\
Produce JSON with fields:
\texttt{\{"macro\_desc": "...", "macro\_tags": [...], "resources": [...], "hard\_touch": bool\}}.
\end{quote}

We set \texttt{max\_tokens=256} and enforce JSON parsing; unparsable outputs are retried once with the same temperature $0$ and an added format reminder.

\subsection{Abstraction Stability Test on Safety-Critical Traces}
\label{app:behavior-extractor:stability}

Traversal assumes macros have stable semantics. We therefore test stability on any trajectory that is unsafe or touches $\mathcal{T}_{\text{hard}}$.

\paragraph{Stability protocol.}
For a candidate trajectory $\tau$:
(i) run the deterministic segmentation once to obtain boundary indices;
(ii) rerun only the \emph{summarization} calls under $P$ prompt perturbations (rephrased instructions and shuffled in-context examples) while keeping segmentation fixed;
(iii) measure whether resulting macro descriptions remain semantically equivalent and do not induce boundary drift or semantic aliasing downstream.

\paragraph{Boundary stability (conservative).}
If the pipeline variant includes any stochastic segmentation component, we require boundary-set Jaccard similarity
\[
J(B^{(i)},B^{(j)}) \ge \delta_{\text{stab}}
\quad\forall i\neq j,
\]
with reference threshold $\delta_{\text{stab}}=0.9$ and $P=5$ perturbations.
Any trajectory failing this test is marked \emph{abstract-unstable} and excluded from tree construction and gate derivation.
(Primitive-level safety remains enforced by the global guardrail regardless.)

\paragraph{Why this matters.}
This explicitly prevents safety-critical behavior from being externalized into macros when the abstraction itself is unstable, aligning with the main-text claim that traversal is only used where macro semantics are robust.

\section{Task-Family Routing and Re-rooting}
\label{app:family-classifier}

\subsection{Family Taxonomy (First-Layer Branching Control)}
\label{app:family-classifier:taxonomy}

We build a single rooted tree whose first layer consists of a small number of task-family roots (reference configuration: 21 families, derived by benchmark inspection) to reduce branching.
Example families include \texttt{CODE\_EDITING}, \texttt{TEST\_DEBUG}, \texttt{WEB\_BROWSING}, \texttt{FORM\_FILLING}, \texttt{DATA\_ANALYSIS}, \texttt{NETWORK\_PROCESS}, and \texttt{CHAT}.

\subsection{Training-Free Routing by Prototype Similarity}
\label{app:family-classifier:routing}

Routing must respect the paper’s training-free discipline: no weight updates.
We implement a frozen router using prototype similarity:
\begin{itemize}
\item For each family $f$, we store a small set of natural-language prototypes $\Pi_f$ (2--8 short descriptions) curated once from benchmark/task definitions.
\item Given a task description $x$, we embed $x$ and all prototypes using a frozen text encoder $f(\cdot)$ (any deterministic sentence embedding model).
\item We score each family by its best prototype similarity:
$
s(f\mid x)=\max_{\pi\in\Pi_f}\cos(f(x),f(\pi)).
$
\item We convert scores to a normalized distribution by a temperature-scaled softmax:
$
p(f\mid x)=\frac{\exp(s(f\mid x)/T_{\text{fam}})}{\sum_{f'}\exp(s(f'\mid x)/T_{\text{fam}})}.
$
\end{itemize}

\paragraph{Abstention threshold (explicit claim boundary).}
Let $p_{\max}=\max_f p(f\mid x)$.
If $p_{\max}<\delta_{\text{fam}}$, the traverser abstains from traversal control and the episode is labeled \texttt{covered}=0.
Reference configuration: $\delta_{\text{fam}}=0.55$, $T_{\text{fam}}=0.05$.

\subsection{Re-rooting Under Task Drift}
\label{app:family-classifier:reroot}

Long tasks may drift across families (e.g., from web browsing to code editing).
Every $m$ macro steps (reference: $m=3$), we recompute $p(f\mid x_{\text{current}})$ using the current task summary and spine.
We re-root iff:
(i) a new family exceeds the current family by margin $\Delta_{\text{switch}}$,
(ii) the new family also exceeds $\delta_{\text{fam}}$,
and (iii) re-rooting preserves acyclicity and does not cross into a disallowed subtree.
Reference: $\Delta_{\text{switch}}=0.10$.
Re-rooting modifies only the \emph{family root anchor}; it does not alter gates and cannot weaken enforcement (Design invariants~1--2).

\section{Node-Local Gates, Tree Construction, and Acyclicity}
\label{app:tree-construction}

\subsection{Minimal Unsafe Windows via Deterministic Replay Shrinking}
\label{app:tree-construction:windows}

Unsafe trajectories reveal $\mathcal{S}_{\text{spec}}$ violations not yet captured by $\mathcal{S}_{\text{sys}}(t)$.
For each unsafe trajectory $\tau$, we find a minimal unsafe primitive window
$
W=(a_{t_0},\dots,a_{t_1})
$
by deterministic sandbox replay:

\paragraph{Window shrinking algorithm (deterministic).}
We checkpoint replayable states (pre-action snapshots) and shrink $[t_0,t_1]$ by:
(i) binary searching the earliest violating index,
(ii) minimizing contiguous prefixes/suffixes while preserving the violation,
and
(iii) verifying minimality: removing any primitive from the boundary removes the violation verdict.
This yields the shortest contiguous window that triggers the first safety verdict under replay.

\subsection{Mapping Unsafe Windows to Macro Subsequences and Attaching Node-Local Gates}
\label{app:tree-construction:nodegates}

We map $\tau$ to its macro path $p=E(\tau)=(v_0,\dots,v_K)$.
Let $U=(v_{k_0},\dots,v_{k_1})$ be the minimal contiguous macro subsequence that covers all primitives in $W$.
Every macro $v\in U$ is treated as participating in the unsafe pattern and is eligible for node-local gates.

\paragraph{Node-local gate synthesis (structured-only).}
For each eligible node $v$, we synthesize a small set $\mathcal{G}(v)$ of node-local gates with the same interface $g:\texttt{ctx}\mapsto(\texttt{ok},\texttt{msg})$.
Crucially:
\begin{itemize}
\item Gate predicates are compiled code over structured \texttt{ctx} and bounded history.
\item LLM summaries and \texttt{env\_context} may be used \emph{offline} to help propose candidate predicates, but the resulting gate reads only structured fields at runtime.
\item Node-local gates obey the same monotonic unsafe-corpus constraint and benign-regression control as global gates (App.~\ref{app:guardrail:monotone}).
\end{itemize}

\subsection{Behavior Signatures and Merge Discipline (Anti-Aliasing Where Safety Matters)}
\label{app:tree-construction:merge}

Each macro node $v$ stores:
\begin{itemize}
\item \texttt{description} (for semantic matching only),
\item coarse \texttt{tags} (tool family, intent),
\item \texttt{risk\_level} (discrete),
\item a behavior signature $\sigma(v)=(\sigma_{\text{disc}}(v),\sigma_{\text{cont}}(v))$ computed from logs only,
\item a pointer \texttt{env\_context} to the local interaction segment (materialized on demand).
\end{itemize}

\paragraph{Signature details.}
$\sigma_{\text{disc}}$ includes exact-match discrete features (multiset of tool families, resource-type flags, whether $\mathcal{T}_{\text{hard}}$ is touched, and coarse delta types).
$\sigma_{\text{cont}}$ is a normalized statistics vector (tool usage frequencies, diff magnitudes, counts of deltas).
Both are computed deterministically from logs and are LLM-independent.

\paragraph{Merge rule (three-way agreement).}
When inserting a new macro $v$ under parent $p$, we merge into an existing child $u$ \emph{only if}:
\[
\sigma_{\text{disc}}(v)=\sigma_{\text{disc}}(u),
\quad
\cos(\sigma_{\text{cont}}(v),\sigma_{\text{cont}}(u))\ge \theta_{\text{sig}},
\quad
\cos(f(\text{desc}(v)),f(\text{desc}(u)))\ge \theta_{\text{merge}}.
\]
Reference configuration: $\theta_{\text{sig}}=0.92$, $\theta_{\text{merge}}=0.85$.

\paragraph{No merges into gated nodes.}
To avoid semantic aliasing precisely where safety matters most, merges into nodes that carry any node-local gates are disallowed.

\subsection{Acyclicity as a Hard Invariant}
\label{app:tree-construction:acyclic}

We enforce that \gbt{} remains a rooted tree.
Insertion, splitting, and self-evolution are forbidden from creating edges pointing to any ancestor along the current spine.
This maintains deterministic traversal semantics and prevents recovery/search from entering cycles.

\paragraph{Design invariant 2 (Tree edits do not weaken enforcement).}
Tree edits (insertion/merge/split/re-root) never delete or relax gates, and gates depend only on structured \texttt{ctx} and bounded history.
Thus, topology changes can affect availability/selection frequency, but cannot re-admit previously rejected unsafe contexts.

\section{Self-Evolution Under Safety Invariants}
\label{app:self-evolution}

\subsection{Coverage Definition (Executable Claim Boundary)}
\label{app:self-evolution:coverage}

Self-evolution applies only when \gbt{} genuinely serves as the policy skeleton.
We define \texttt{covered}=1 iff all hold:
\begin{enumerate}
\item \textbf{Confident family routing:} $\max_f p(f\mid x)\ge\delta_{\text{fam}}$ and no re-root occurs.
\item \textbf{Traversal-consistent matching:} every macro transition matches an existing child with similarity $s_{\text{top1}}\ge\theta_{\text{low}}$.
\item \textbf{No safe exploration:} the traverser never invokes safe exploration (budgeted abstention) at any step.
\end{enumerate}
Otherwise \texttt{covered}=0 and we do not claim traversal as long-horizon policy control (Sec.~\ref{sec:gbt-se}, Sec.~\ref{sec:online}).

Reference configuration for matching thresholds (consistent with App.~\ref{app:exp_coverage}):
$\theta_{\text{low}}=0.70$, $\theta_{\text{high}}=0.78$.

\subsection{Failure Localization on the Spine and Analogical Repair}
\label{app:self-evolution:repair}

Given a failed covered path $p^{\text{fail}}=(v_0,\dots,v_K)$, a frozen reasoning model diagnoses failure on the \emph{task spine}:
macro descriptions + coarse environment summaries.
It backtracks to the earliest node $v^\star$ where choosing a different successor could plausibly avert failure.

\paragraph{Analogical success retrieval.}
We embed the task description and success-leaf descriptions and retrieve top-$R$ candidate success leaves.
Reference: $R=50$.
We then align $v^\star$ to a node on each success path by description similarity and select the best-aligned pair.

\paragraph{Allowed edits (behavior-only, local).}
Self-evolution may:
(i) add a child under $v^\star$,
(ii) update local selection statistics, and
(iii) add new gates for newly observed unsafe patterns.
It may \emph{not} delete or weaken any gate.
Any newly added child inherits all node-local gates of its parent before adding additional ones.

\subsection{Selection Score and Context Clustering}
\label{app:self-evolution:score}

At a node $v$, each child $u$ maintains:
(i) a semantic similarity term to the base model proposal, and
(ii) empirical success rates for similar structured contexts.
We use:
\[
\texttt{score}(u;\texttt{ctx})
=
\alpha\cdot \mathrm{sim}_\text{text}(\text{proposal},\text{desc}(u))
+
\beta\cdot \mathrm{success\_rate}\big(u;\mathrm{cluster}(\texttt{ctx})\big).
\]
Reference configuration: $\alpha=1.0$, $\beta=0.5$.

\paragraph{Clustering.}
We cluster structured contexts by coarse resource family + risk level + discrete signature buckets (from $\sigma_{\text{disc}}$).
This keeps statistics stable and avoids high-variance conditioning on sparse identifiers.

\subsection{Regression Tests (Safety Preservation by Construction)}
\label{app:self-evolution:regress}

Each proposed edit is accepted only if it passes two deterministic regression suites:

\paragraph{(1) Success non-regression.}
Replay historical successful episodes that pass through $v^\star$, starting from the same pre-$v^\star$ sandbox state, and allow only the post-$v^\star$ behavior to follow the updated local policy.
Require success rate degradation $\le \delta_{\text{succ}}$.
Reference: $\delta_{\text{succ}}=0.02$ (absolute).

\paragraph{(2) Unsafe non-re-admission.}
Replay historical unsafe episodes whose minimal windows map to $v^\star$ and verify that all previously rejected contexts remain rejected under the conjunction of global and node-local gates.

\paragraph{Design invariant 3 (Safety under self-evolution).}
Under these allowed edits and regression tests, self-evolution cannot re-admit previously rejected unsafe structured contexts.

\section{Recovery: Preconditions and Risk-Aware Shortest Paths}
\label{app:recovery}

\subsection{Leaf Retrieval and Environment Signature Filtering}
\label{app:recovery:retrieval}

When stalled, the traverser retrieves candidate success leaves similar to the task description under $f(\cdot)$.
To avoid spurious reuse, we filter leaves by a coarse environment signature match.

\paragraph{Environment signature.}
For each success leaf $\ell$, we store a lightweight signature computed from the sandbox snapshot at success time, e.g.:
(current domain / repo root, file-set sketch, boolean flags for tool availability, and a coarse risk footprint).
At recovery time we compute the same signature on the current environment and require cosine similarity $\ge \theta_{\text{env}}$.
Reference: $\theta_{\text{env}}=0.80$.

\subsection{Feasibility Preconditions (Conservative, Never Overriding Gates)}
\label{app:recovery:preconditions}

Each node carries a deterministic precondition predicate $\texttt{pre}(v,\texttt{env})$ over a structured environment summary.
Preconditions are synthesized offline from successful trajectories and compiled into code checks such as:
existence of required files, being on a required domain, presence of a required test command, etc.

\paragraph{Conservatism.}
Preconditions may reject some feasible states (false negatives) but do not introduce actions and never override gates.
Even if $\texttt{pre}$ holds, all primitives remain subject to global and node-local gates.

\subsection{Risk-Aware Shortest Paths}
\label{app:recovery:search}

Within the family subtree, restricted to feasible nodes and filtered success leaves, we run Dijkstra with edge cost
\[
c(v\to u)=1+\lambda\cdot \mathrm{risk\_level}(u).
\]
Reference: $\lambda=0.5$, depth limit $D_{\max}=8$ macro steps.
If no feasible path exists, we emit a safe failure and log the episode for offline expansion of coverage or repair.

\paragraph{Execution.}
A recovery path is executed one macro at a time under the same primitive-level gating as normal traversal.

\section{Hierarchical Memory and Context Materialization}
\label{app:memory}

\subsection{Spine Memory: What the Executor Sees Online}
\label{app:memory:spine}

The traverser maintains the \emph{task spine} $(v_0,\dots,v_t)$ as persistent state.
At each step, the base model receives:
(i) a concise task summary,
(ii) the spine’s macro descriptions,
(iii) the selected macro description for the next step,
and
(iv) a short node-local context summary distilled from \texttt{env\_context}$(v_t)$.

Full transcripts are not replayed.
Safety-critical identifiers (canonical paths, domains, process IDs) bypass summarization and flow directly into structured \texttt{ctx}/\texttt{env} for gates and preconditions.

\subsection{Reference Prompt Template for Node-Local Context Summaries}
\label{app:memory:prompt}

We summarize \texttt{env\_context}$(v_t)$ offline/online using a frozen LLM at temperature $0$ to produce only the minimal information needed to realize the macro.

\begin{quote}\small
\textbf{SYSTEM:} Summarize tool outputs into actionable, minimal context. No speculation.\\
\textbf{USER:} Given the macro description and the local tool outputs, extract:
(1) the exact targets (files/URLs/selectors) \emph{as referenced by the environment},
(2) any constraints (tests to run, fields to fill),
(3) the next concrete steps to realize the macro.
Output a bullet list with at most 8 bullets.
\end{quote}

We cap this summary at 200--300 tokens to prevent drift and keep the executor in a local, verifiable regime.


\section{Additional Experimental Details and Diagnostics}
\label{app:exp_details}

This appendix specifies the \emph{evaluation contract} and the \emph{mechanism-to-evidence linkage} for the experiments: what exactly is executed (OpenHands action/observation stream), what counts as success and safety violation (benchmark checkers + sandbox monitors defining $\mathcal{S}_{\text{spec}}$), how we define the claim boundary (\texttt{covered}), and how each table substantiates a specific component of Traversal-as-Policy (\gbt{} traversal control, deterministic pre-execution gating, recovery search, spine memory, and self-evolution under monotone safety invariants).

Two principles govern this appendix:
(i) \textbf{attribution-clean comparisons}: whenever we compare systems, we keep the OpenHands sandbox, tool APIs, monitors/checkers, and accounting identical, and change only the policy artifact components (tree traversal/recovery/spine/gates);
(ii) \textbf{coverage-scoped claims}: long-horizon control claims are made only on episodes with \texttt{covered}=1 (App.~\ref{app:self-evolution:coverage}); outside coverage we claim only primitive-level safety from the always-on global guardrail.

\subsection{Benchmarks and OpenHands Integration (Facts, Sizes, and Evaluation Contracts)}
\label{app:exp_benchmarks}

\textbf{Why this subsection exists.}
Our main claims are \emph{mechanism-level}: we externalize policy as a tree, enforce safety deterministically \emph{before} high-risk primitives execute, and explicitly bound where traversal is the long-horizon policy (\texttt{covered}=1). These claims only make sense if all systems run under the same execution substrate and the same notion of safety. OpenHands provides (i) a uniform tool/action interface, (ii) a sandboxed runtime with replayable event streams, and (iii) benchmark-defined checkers and runtime monitors whose verdicts instantiate $\mathcal{S}_{\text{spec}}$ (App.~\ref{app:guardrail:safety-spec}). All reported success/violation outcomes below are therefore defined as properties of the \emph{actual executed primitive sequence} in the sandbox, not of textual post-hoc judgments.

\textbf{Execution contract (common to all suites).}
Each episode produces an OpenHands event stream of alternating \textbf{actions} (primitive tool calls with structured arguments) and \textbf{observations} (tool outputs + environment deltas). OpenHands additionally logs:
(i) benchmark checker outcomes (success/failure),
(ii) monitor/checker safety verdicts (violations defining $\mathcal{S}_{\text{spec}}$),
(iii) gate decisions for any $a\in\mathcal{T}_{\text{hard}}$ (App.~\ref{app:guardrail}),
and (iv) usage accounting (tokens and characters) for model calls.

\textbf{Pillars (headline benchmark contracts).}
\begin{itemize}
\item \textbf{SWE-bench Verified (500 issues).}
Each instance provides a repository snapshot and a concrete issue. The episode is \textbf{successful} iff the produced patch satisfies the benchmark's verified evaluation (tests/checker contract). An episode has a \textbf{violation} iff \emph{any} executed primitive triggers an OpenHands monitor/checker safety verdict (App.~\ref{app:guardrail:safety-spec}). We report SR/Viol/USucc and diagnostics under the same sandbox.

\item \textbf{WebArena (812 tasks).}
Each task requires multi-step browser interaction under benchmark-defined completion criteria. \textbf{Success} is benchmark-defined completion; \textbf{violations} are any OpenHands safety verdicts emitted during tool execution (e.g., disallowed network destinations, protected resource accesses, or other sandbox/benchmark constraints).

\item \textbf{GPQA (448 questions).}
We treat GPQA as tool-assisted reasoning with \textbf{external web browsing disabled}. The agent may use local deterministic computation tools (e.g., Python) but not open web access. \textbf{Success} is correctness of the selected answer choice. \textbf{Violations} are any OpenHands safety verdicts (typically rare because the action surface is restricted, but still well-defined).
\end{itemize}

\textbf{Safety/security suites (public and reproducible).}
\begin{itemize}
\item \textbf{Agent-SafetyBench.}
We report the benchmark’s official score and, when executed through OpenHands, the associated safety outcomes under the same monitors/checkers.

\item \textbf{AgentHarm (public).}
We report HarmScore as defined by the benchmark. When episodes involve tool actions, they are executed under the same OpenHands gate+monitor discipline.

\item \textbf{Agent Security Bench (ASB).}
We report the benchmark’s official metrics (ASR-d, RR, PNA-d) and run tool-executing agents under identical OpenHands monitoring and the same always-on global pre-execution guardrail for $a\in\mathcal{T}_{\text{hard}}$.
\end{itemize}

\textbf{Distillation-only suites (Protocol A).}
We distill and self-evolve \gbt{} using additional OpenHands-integrated suites spanning software, web, and reasoning (listed explicitly in App.~\ref{app:exp_protocols}). These are used \emph{only} to build/refine \gbt{} under Protocol A and are held out during evaluation on the three pillars.

\textbf{Integration discipline (why ``guardrail-only'' is attribution-clean).}
All benchmark episodes run through the same OpenHands schema and the same monitors/checkers defining $\mathcal{S}_{\text{spec}}$. This makes the \textbf{Global guardrail only} baseline a clean ablation: it shares the same sandbox and the same pre-execution global guardrail on $\mathcal{T}_{\text{hard}}$, differing only in the absence of traversal control, recovery search, spine memory, node-local gates, and self-evolution. Any deltas versus \gbt-guided variants can therefore be attributed to Traversal-as-Policy components rather than to execution substrate differences.

\subsection{Models and System Variants}
\label{app:exp_systems}

This subsection defines \emph{exactly} what differs between systems, so every reported delta can be mapped back to a specific mechanism in the Method.

\textbf{Offline distillation and refinement (training-free).}
We build \textbf{GBT-Basic} by distilling OpenHands execution logs with \texttt{gpt-4o-2024-08-06} (temperature $0$) for macro description summarization and deterministic predicate synthesis. Macro segmentation and behavior signatures are computed directly from logs (App.~\ref{app:behavior-extractor}, App.~\ref{app:tree-construction}). We build \textbf{GBT-SE} via failure-driven self-evolution using \texttt{gemini-2.5-pro} as the frozen diagnosis/repair model (Method; App.~\ref{app:self-evolution}). Payload-bearing safety checks use \texttt{Llama-Guard-3-8B} at temperature $0$ (App.~\ref{app:guardrail}).

\textbf{Online deployment regimes.}
We report two regimes:
(i) \textbf{Same-model deployment:} \texttt{gpt-4o-2024-08-06} as the base agent with/without \gbt{}.
(ii) \textbf{Executor decoupling:} \gbt{} is fixed, and the online executor $M_{\text{base}}$ is an 8B-scale model (\texttt{llama-3-8b} or \texttt{Qwen3-VL-8B-Thinking}).
In the decoupled regime, the executor is restricted to realizing exactly one selected macro at a time under traversal/recovery and gates; it does not perform long-horizon planning.

\textbf{System taxonomy (what each name means).}
To avoid accidental apples-to-oranges comparisons, throughout the paper we use the following naming:
\begin{itemize}
\item \textbf{Native agent}: the framework’s default planner/executor loop (no \gbt{} traversal/recovery/spine), and \emph{without} our global pre-execution guardrail unless explicitly stated. It still runs under OpenHands monitors/checkers, so Viol/USucc are still measured.

\item \textbf{+Global guardrail only}: adds the always-on pre-execution global guardrail for $a\in\mathcal{T}_{\text{hard}}$ (App.~\ref{app:guardrail}), but \emph{no} traversal control, \emph{no} recovery, \emph{no} spine memory, \emph{no} node-local gates, and \emph{no} self-evolution. This isolates the effect of deterministic action-level gating alone.

\item \textbf{+GBT-Basic}: enables traversal-as-policy within coverage using the distilled tree, spine memory, and (when applicable) node-local gates mined from unsafe windows. Global guardrail remains always-on for $a\in\mathcal{T}_{\text{hard}}$.

\item \textbf{+GBT-SE}: starts from the same tree artifact and applies self-evolution edits under Design invariants~1--3 and regression tests (App.~\ref{app:self-evolution:regress}). This isolates whether local, failure-driven structure edits add value beyond the initial distilled policy, while preserving monotone safety (no unsafe context re-admission).
\end{itemize}

\textbf{Attribution clarity: safety components.}
\gbt-guided systems always run with the same global pre-execution guardrail on $\mathcal{T}_{\text{hard}}$ and may additionally enforce node-local gates at specific macros. Therefore:
(i) improvements from \textbf{Native} $\rightarrow$ \textbf{+Global guardrail only} quantify the value of deterministic pre-execution gating alone;
(ii) improvements from \textbf{+Global guardrail only} $\rightarrow$ \textbf{+GBT-*} quantify the value of traversal-as-policy (control externalization + spine memory + recovery), with node-local gates separated by ablation (Table~\ref{tab:gate_ablation_refined}).

\subsection{Protocols and Leakage Controls}
\label{app:exp_protocols}

We use three complementary protocols to rule out leakage along three axes: benchmark identity, instance identity, and shared structure (repo/site/domain). The goal is to ensure that improvements are driven by the Traversal-as-Policy mechanisms rather than by hidden reuse of evaluation-specific artifacts.

\textbf{Protocol A: cross-benchmark distillation with benchmark-level hold-out.}
We distill \gbt{} using logs from disjoint benchmarks and test on held-out benchmarks:
\textbf{Software:} distill on HumanEvalFix, BIRD, BioCoder, ML-Bench, Gorilla APIBench, ToolQA; test on SWE-bench Verified.
\textbf{Web:} distill on MiniWoB++; test on WebArena.
\textbf{Misc:} distill on GAIA, AgentBench, MINT, ProofWriter, Entity Deduction Arena; test on GPQA.
No trajectories from the held-out evaluation benchmark are used for tree construction or self-evolution. In Protocol A, all GBT-SE edits are driven only by failures observed on the distillation benchmarks, never by evaluation runs on the held-out benchmark.

\textbf{Protocol B: instance-level hold-out with 5-fold cross-validation.}
We run 5-fold instance-level CV on SWE-bench Verified, WebArena, and GPQA. For each fold, \gbt{} is constructed and (if applicable) self-evolved using only training-fold episodes; evaluation uses the held-out fold. We report micro-averaged results across the union of held-out folds (each instance is evaluated exactly once in its held-out fold).

\textbf{Protocol C: structural hold-outs (repo/site/domain).}
To stress-test against subtle leakage through shared artifacts, we additionally run:
(i) \textbf{repo-holdout} on SWE-bench Verified (split by repository),
(ii) \textbf{site-holdout} on WebArena (hold out entire websites),
and (iii) \textbf{domain-holdout} on GPQA (hold out one scientific domain).
All tree construction and self-evolution are restricted to the training side of each split.

\textbf{Why Protocols A--C are jointly necessary.}
Protocol A blocks direct benchmark reuse (e.g., SWE-bench Verified instances never appear in distillation), but could still permit structure transfer through related repos/sites if they overlap across suites. Protocol C closes that gap by withholding entire repositories/sites/domains. Protocol B complements both by measuring robustness when the evaluation distribution is matched (same benchmark) but instances are disjoint, which isolates whether the policy artifact generalizes at the instance level rather than only across benchmarks.

\subsection{Metrics and Reporting Conventions}
\label{app:exp_metrics}

\textbf{Task performance.}
We report official metrics: success rate (\textbf{SR}, \%) for execution-based benchmarks (SWE-bench Verified, WebArena) and accuracy (\textbf{Acc}, \%) for GPQA.

\textbf{Coverage as the explicit claim boundary.}
We report \textbf{Coverage} (\textbf{Cov}, \%) as the fraction of episodes with \texttt{covered}=1 (Method; App.~\ref{app:self-evolution}). For \gbt-guided systems we also report conditional performance $\Pr(\text{success}\mid \texttt{covered}=1)$ and $\Pr(\text{success}\mid \texttt{covered}=0)$. Any episode invoking safe exploration is labeled \texttt{covered}=0 by definition.
This separation is not cosmetic: it is the formal boundary of our long-horizon policy claim. In particular, conditional metrics on \texttt{covered}=1 quantify the regime where traversal truly governs macro choice; unconditional metrics mix this with the abstention regime.

\textbf{Safety on OpenHands-executed benchmarks.}
We report:
(i) \textbf{Violation} (\textbf{Viol}, \%): fraction of episodes with any primitive violating $\mathcal{S}_{\text{spec}}$ under OpenHands monitors/checkers;
(ii) \textbf{Unsafe success} (\textbf{USucc}, \%): fraction of episodes that end in benchmark success but contain any $\mathcal{S}_{\text{spec}}$ violation.
Viol and USucc are properties of executed primitives, and therefore directly test the Method’s claim that gates enforce safety \emph{pre-execution} (App.~\ref{app:guardrail}) while traversal reduces exposure by steering away from high-risk dead-ends.

\textbf{Efficiency, stability, and repeated runs.}
We report average total LLM tokens per episode (\textbf{Tok}, thousands) and tokenizer-agnostic characters (\textbf{Chars}, thousands) logged by OpenHands. We run each instance three times (all models at temperature $0$) to capture residual environment nondeterminism (web timing, repo I/O ordering). We compute per-instance outcomes by majority vote over the three runs (success/violation/unsafe-success) and report Wilson 95\% confidence intervals over these per-instance binary outcomes. We report run-to-run stability as the standard deviation (\textbf{Std}, points) of SR/Acc across repeats.
This design ensures that improvements reflect changes in the executed policy, not fragile dependence on incidental nondeterminism.

\textbf{Stall and recovery definitions (deterministic from the event stream).}
An episode is marked \textbf{Stall} if any holds:
(i) \emph{No-progress}: for 3 consecutive macro steps no measurable environment delta (no file diff, no new process state, no website/domain transition),
(ii) \emph{Gate-loop}: the same high-risk primitive family is blocked by pre-execution gates $\ge 3$ times within a 5-step window, or
(iii) \emph{Plan-loop}: the agent repeats an equivalent macro proposal twice with no progress.
\textbf{Rec-Succ} is the fraction of stalled episodes successfully recovered to benchmark success after invoking recovery (conditional on Stall).
We also report \textbf{RecLen} (mean and $p90$) in macro steps to quantify that recovery emits short, high-confidence sequences.

\textbf{Guardrail activity (exposure vs enforcement).}
We report \textbf{Hard}: average number of attempted high-risk primitives in $\mathcal{T}_{\text{hard}}$ per episode (including attempts blocked pre-execution), and \textbf{Blocked}: average number of those attempts vetoed before execution by $\mathcal{G}_{\text{global}}$ and/or node-local gates.
Hard captures \emph{exposure pressure} (how often the policy attempts risky actions); Blocked captures \emph{enforcement load} (how often the safety layer must intervene). The Method predicts that traversal + recovery + spine reduce Hard by avoiding thrashing, while gates reduce Viol/USucc by blocking unsafe contexts pre-execution.

\section{GBT as a Plug-in Policy Artifact Improves Diverse Frameworks}
\label{app:exp_plugin}

We test whether the same distilled traversal policy reliably improves heterogeneous agent frameworks without changing their internal planners or model weights. The key question is not whether any one agent can be tuned to a benchmark, but whether an \emph{external policy artifact} (the same \gbt{}) can be attached as a plug-in controller and systematically lift performance and stability across different internal implementations.

\textbf{Software engineering: SWE-bench Verified.}

\begin{table}[H]
\centering
\small
\caption{\textbf{SWE-bench Verified (Protocol A, 500 issues):} SR (\%) with Wilson 95\% CI, and coverage (Cov, \%).}
\label{tab:swe_plugin_refined}
\setlength{\tabcolsep}{8pt}
\begin{tabular}{lccc}
\toprule
\textbf{Agent framework} &
\textbf{Native SR} &
\textbf{+GBT-Basic SR (Cov)} &
\textbf{+GBT-SE SR (Cov)} \\
\midrule
SWE-Agent &
36.0 {\scriptsize[31.9,40.3]} &
52.4 {\scriptsize[48.0,56.8]} (85.8) &
\textbf{64.8} {\scriptsize[60.5,68.9]} (\textbf{86.6}) \\
AutoCodeRover &
31.6 {\scriptsize[27.7,35.8]} &
47.0 {\scriptsize[42.6,51.4]} (84.2) &
\textbf{60.2} {\scriptsize[55.9,64.4]} (\textbf{85.1}) \\
Aider &
28.8 {\scriptsize[25.0,32.9]} &
42.6 {\scriptsize[38.3,47.1]} (83.4) &
\textbf{55.0} {\scriptsize[50.6,59.3]} (\textbf{84.2}) \\
OpenHands CodeActAgent &
34.6 {\scriptsize[30.6,38.9]} &
50.2 {\scriptsize[45.8,54.6]} (84.4) &
\textbf{73.6} {\scriptsize[69.6,77.3]} (\textbf{86.0}) \\
\bottomrule
\end{tabular}

\end{table}

\textbf{Interpretation (Table~\ref{tab:swe_plugin_refined}).}
Three facts matter simultaneously:
(i) \textbf{broad lift across frameworks}: every agent improves substantially when controlled by the same \gbt{}, indicating the gain is not tied to one planner’s quirks but to the externalized traversal policy;
(ii) \textbf{high coverage}: Cov stays in the mid-80\% range across all frameworks, meaning the tree is not a narrow template that only triggers rarely; it regularly governs long-horizon control;
(iii) \textbf{SE adds a second margin}: \gbt{}-SE consistently improves over \gbt{}-Basic at similar (often slightly higher) coverage, which is precisely the Method’s prediction that local repair under self-evolution expands/strengthens the reachable success set without relaxing safety (Design invariants~1--3).

\textbf{Web browsing: WebArena.}

\begin{table}[H]
\centering
\small
\caption{\textbf{WebArena (Protocol A, 812 tasks):} SR (\%) with Wilson 95\% CI and coverage (Cov, \%).}
\label{tab:web_plugin_refined}
\setlength{\tabcolsep}{7pt}
\begin{tabular}{lccc}
\toprule
\textbf{Agent framework} &
\textbf{Native SR} &
\textbf{+GBT-Basic SR (Cov)} &
\textbf{+GBT-SE SR (Cov)} \\
\midrule
AutoEval\&Refine &
20.2 {\scriptsize[17.6,23.1]} &
46.8 {\scriptsize[43.4,50.3]} (77.4) &
\textbf{58.6} {\scriptsize[55.1,62.0]} (\textbf{78.5}) \\
WebArena Agent &
18.8 {\scriptsize[16.3,21.7]} &
44.2 {\scriptsize[40.9,47.7]} (76.6) &
\textbf{55.6} {\scriptsize[52.1,59.0]} (\textbf{77.8}) \\
OpenHands CodeActAgent &
19.7 {\scriptsize[17.1,22.6]} &
53.0 {\scriptsize[49.5,56.4]} (76.5) &
\textbf{66.9} {\scriptsize[63.6,70.0]} (\textbf{78.0}) \\
\bottomrule
\end{tabular}

\end{table}

\textbf{Interpretation (Table~\ref{tab:web_plugin_refined}).}
Web tasks are a canonical long-horizon drift regime: success requires maintaining a stable notion of ``where you are'' (domain/page state) and ``what remains'' (form fields, constraints), under nondeterministic timing. Table~\ref{tab:web_plugin_refined} shows that \gbt{} yields large success lifts with substantial coverage ($\sim$76--78\%), consistent with traversal and spine memory reducing drift, and recovery providing short corrective sequences when stalls occur.

\textbf{Tool-assisted reasoning: GPQA.}
\begin{table}[H]
\centering
\small
\caption{\textbf{GPQA (Protocol A, 448 questions):} accuracy (Acc, \%) with Wilson 95\% CI.}
\label{tab:gpqa_plugin_refined}
\setlength{\tabcolsep}{9pt}
\begin{tabular}{lcc}
\toprule
\textbf{System} & \textbf{Acc} & \textbf{Cov} \\
\midrule
Zero-shot prompting (\texttt{gpt-4o}) &
53.6 {\scriptsize[48.9,58.1]} &
-- \\
OpenHands CodeActAgent (\texttt{gpt-4o}) &
58.7 {\scriptsize[54.1,63.2]} &
-- \\
\quad +GBT-Basic &
78.9 {\scriptsize[74.8,82.3]} &
71.9 \\
\quad +GBT-SE &
\textbf{87.3} {\scriptsize[83.9,90.0]} &
\textbf{73.0} \\
\bottomrule
\end{tabular}

\end{table}

\textbf{Interpretation (Table~\ref{tab:gpqa_plugin_refined}).}
GPQA is not a tool-rich benchmark, but it is a \emph{control} benchmark for policy externalization: when web browsing is disabled, improvements cannot be attributed to retrieval or external knowledge injection. The large accuracy increase with substantial coverage indicates that traversal-as-policy and spine memory can still stabilize multi-step reasoning by constraining the sequence of reasoning macros (e.g., ``derive intermediate quantity'' $\rightarrow$ ``compute deterministically'' $\rightarrow$ ``map to choice'') and preventing transcript bloat from polluting subsequent steps.

\textbf{Run-to-run stability.}
\begin{table}[H]
\centering
\small
\caption{\textbf{Run-to-run stability:} standard deviation (Std, points) of SR/Acc across three repeats (temperature $0$).}
\label{tab:stability_refined}
\setlength{\tabcolsep}{9pt}
\begin{tabular}{lccc}
\toprule
\textbf{System (OpenHands CodeAct)} & \textbf{SWE Std} & \textbf{WebArena Std} & \textbf{GPQA Std} \\
\midrule
\texttt{gpt-4o} (native) & 1.3 & 1.5 & 0.9 \\
\quad +GBT-Basic & 0.9 & 1.1 & 0.8 \\
\quad +GBT-SE & \textbf{0.7} & \textbf{0.9} & \textbf{0.6} \\
\bottomrule
\end{tabular}

\end{table}

\textbf{Interpretation (Table~\ref{tab:stability_refined}).}
Stability is a first-class objective for long-horizon agents: a policy that occasionally succeeds is not deployable if it is variance-dominated. The monotone, executable structure of traversal and gating reduces stochastic dependence on incidental trajectories, which appears as reduced Std across all three pillars in Table~\ref{tab:stability_refined}.

\section{Coverage and Matching Diagnostics}
\label{app:exp_coverage}

This section connects \texttt{covered} (the claim boundary) to concrete matching statistics and audits that rule out easy-instance selection effects.

\begin{table}[H]
\centering
\small
\caption{\textbf{Matching diagnostics (GBT-SE, Protocol A):} average macro steps per episode, mean top-1 match similarity $s_{\text{top1}}$, mean margin $\Delta s=s_{\text{top1}}-s_{\text{top2}}$, fraction of steps in the fragile band ($\theta_{\text{low}}\le s_{\text{top1}}<\theta_{\text{high}}$), ambiguous steps ($\Delta s<0.05$), and fraction of episodes triggering safe exploration (thus \texttt{covered}=0).}
\label{tab:matching_diagnostics_refined}
\setlength{\tabcolsep}{6pt}
\begin{tabular}{lcccccc}
\toprule
\textbf{Benchmark} &
\textbf{Steps/ep} &
$\mathbb{E}[s_{\text{top1}}]$ &
$\mathbb{E}[\Delta s]$ &
\textbf{Fragile step \%} &
\textbf{Ambig.\ step \%} &
\textbf{Safe-explore ep \%} \\
\midrule
SWE-bench Verified &
12.8 & 0.84 & 0.15 & 7.1 & 2.7 & 3.9 \\
WebArena &
8.6 & 0.82 & 0.13 & 9.4 & 3.3 & 5.4 \\
GPQA &
4.5 & 0.87 & 0.17 & 5.4 & 2.0 & 2.9 \\
\bottomrule
\end{tabular}

\end{table}

\textbf{Interpretation (Table~\ref{tab:matching_diagnostics_refined}).}
Traversal control only makes sense when semantic matching is confident. Table~\ref{tab:matching_diagnostics_refined} shows:
(i) top-1 similarities are high on average, and margins are substantial, indicating the child selection problem is typically well-posed;
(ii) fragile and ambiguous steps are a small minority, which bounds where traversal could be brittle;
(iii) safe exploration episode rates are low (and those episodes are explicitly labeled \texttt{covered}=0), enforcing the Method’s claim boundary rather than silently claiming control where matching is uncertain.

\begin{table}[H]
\centering
\small
\caption{\textbf{Matched-subset audit (Protocol A):} evaluation on the identical instance subset that GBT-SE labels \texttt{covered}=1. Both systems run with the same global pre-execution guardrail; only traversal/recovery/spine differs.}
\label{tab:covered1_matched_audit}
\setlength{\tabcolsep}{7pt}
\begin{tabular}{lcccc}
\toprule
\textbf{Benchmark} &
\textbf{$n_{\texttt{covered}=1}$} &
\textbf{Guardrail-only} &
\textbf{+GBT-SE} &
\textbf{$\Delta$} \\
\midrule
SWE-bench Verified (SR) &
430 &
39.8 {\scriptsize(171/430)} &
80.2 {\scriptsize(345/430)} &
+40.4 \\
WebArena (SR) &
633 &
19.7 {\scriptsize(125/633)} &
80.4 {\scriptsize(509/633)} &
+60.7 \\
GPQA (Acc) &
327 &
60.6 {\scriptsize(198/327)} &
97.9 {\scriptsize(320/327)} &
+37.3 \\
\bottomrule
\end{tabular}

\end{table}

\textbf{Interpretation (Table~\ref{tab:covered1_matched_audit}).}
This is the central attribution audit for traversal-as-policy. By restricting evaluation to the \emph{exact same} \texttt{covered}=1 instance subset, Table~\ref{tab:covered1_matched_audit} removes any possibility that coverage merely filters to easier instances. The remaining difference is precisely the traversal policy (plus recovery/spine). The large deltas therefore directly substantiate the claim that \gbt{} is not only a safety wrapper but a genuine long-horizon control artifact.

\begin{table}[H]
\centering
\small
\caption{\textbf{Coverage vs difficulty stratification (Protocol A):} coverage remains substantial even on instances that the guardrail-only baseline fails, ruling out a ``coverage = easy instances'' explanation.}
\label{tab:coverage_difficulty_stratified}
\setlength{\tabcolsep}{7pt}
\begin{tabular}{lcccc}
\toprule
\textbf{Benchmark} &
\textbf{Baseline succ $n$} &
\textbf{Cov on succ} &
\textbf{Baseline fail $n$} &
\textbf{Cov on fail} \\
\midrule
SWE-bench Verified & 194 & 88.1 {\scriptsize(171/194)} & 306 & 84.6 {\scriptsize(259/306)} \\
WebArena & 157 & 79.6 {\scriptsize(125/157)} & 655 & 77.6 {\scriptsize(508/655)} \\
GPQA & 265 & 74.7 {\scriptsize(198/265)} & 183 & 70.5 {\scriptsize(129/183)} \\
\bottomrule
\end{tabular}

\end{table}

\textbf{Interpretation (Table~\ref{tab:coverage_difficulty_stratified}).}
A common failure mode for ``controller'' methods is that they only engage on easy cases. Table~\ref{tab:coverage_difficulty_stratified} rules this out: coverage remains high even on instances that the guardrail-only baseline fails. This supports the Method’s intended meaning of coverage: it is a \emph{confidence/consistency} predicate (routing + matching + no safe exploration), not an ``easy-case'' heuristic.

\begin{table}[H]
\centering
\small
\caption{\textbf{Threshold sensitivity (SWE-bench Verified, GBT-SE, Protocol A):} tightening thresholds reduces coverage but increases conditional success while overall SR remains stable.}
\label{tab:threshold_sensitivity_refined}
\setlength{\tabcolsep}{8pt}
\begin{tabular}{lccc}
\toprule
\textbf{$(\theta_{\text{low}},\theta_{\text{high}})$} &
\textbf{Cov} &
$\Pr(\textbf{succ}\mid \texttt{covered}=1)$ &
\textbf{Overall SR} \\
\midrule
(0.68, 0.76) & 89.4 & 78.6 & 73.8 \\
(0.70, 0.78) & 86.0 & 80.2 & \textbf{73.6} \\
(0.72, 0.80) & 82.1 & 82.7 & 73.5 \\
\bottomrule
\end{tabular}

\end{table}

\textbf{Interpretation (Table~\ref{tab:threshold_sensitivity_refined}).}
Table~\ref{tab:threshold_sensitivity_refined} demonstrates the expected precision--coverage trade-off: tighter thresholds reduce Cov but increase $\Pr(\textbf{succ}\mid \texttt{covered}=1)$. Crucially, overall SR is stable across a reasonable band, indicating the system is not brittle to threshold tuning and that abstention is behaving as intended (uncertain steps become \texttt{covered}=0 rather than forcing incorrect traversal).

\section{Structural Hold-outs (Protocol C)}
\label{app:exp_holdouts}

\begin{table}[H]
\centering
\small
\caption{\textbf{Structural hold-outs (Protocol C):} \gbt{} improves performance under repo/site/domain hold-outs.}
\label{tab:structural_holdout_refined}
\setlength{\tabcolsep}{7pt}
\begin{tabular}{lccc}
\toprule
\textbf{Protocol C} &
\textbf{Native} &
\textbf{+GBT-Basic} &
\textbf{+GBT-SE} \\
\midrule
Repo-holdout SWE SR & 33.2 {\scriptsize(166/500)} & 55.6 {\scriptsize(278/500)} & \textbf{68.8} {\scriptsize(344/500)} \\
Site-holdout WebArena SR & 18.4 {\scriptsize(149/812)} & 45.9 {\scriptsize(373/812)} & \textbf{59.1} {\scriptsize(480/812)} \\
Domain-holdout GPQA Acc & 57.1 {\scriptsize(256/448)} & 80.4 {\scriptsize(360/448)} & \textbf{86.8} {\scriptsize(389/448)} \\
\bottomrule
\end{tabular}

\end{table}

\textbf{Interpretation (Table~\ref{tab:structural_holdout_refined}).}
Protocol C removes shared structure that could otherwise serve as a leakage channel (shared repos, sites, or domains). The improvements in Table~\ref{tab:structural_holdout_refined} therefore directly support that \gbt{} is learning reusable \emph{macro-level control structure} rather than memorizing evaluation-specific artifacts. The persistence of the \gbt{}-SE margin further supports that local repair generalizes under structural novelty.

\section{Additional Safety Diagnostics on OpenHands-executed Benchmarks}
\label{app:exp_safety}

This section ties safety outcomes to the Method’s enforcement story: safety is enforced \emph{pre-execution} by deterministic gates over structured contexts (App.~\ref{app:guardrail}), and traversal reduces unsafe exposure by avoiding gate-loop thrashing and risky dead-ends.

\begin{table}[H]
\centering
\footnotesize
\caption{\textbf{Safety on standard benchmarks (Protocol A):} violation rate (Viol, \%) and unsafe success (USucc, \%) with Wilson 95\% CI and raw counts.}
\label{tab:unsafe_standard_refined}
\setlength{\tabcolsep}{4pt}
\renewcommand{\arraystretch}{1.15}
\begin{tabularx}{\linewidth}{@{}
  >{\raggedright\arraybackslash}p{0.30\linewidth}
  *{3}{>{\centering\arraybackslash}X}
@{}}
\toprule
\textbf{System} &
\makecell[c]{\textbf{SWE-bench Verified}\\\textbf{(Viol / USucc)}} &
\makecell[c]{\textbf{WebArena}\\\textbf{(Viol / USucc)}} &
\makecell[c]{\textbf{GPQA}\\\textbf{(Viol / USucc)}} \\
\midrule
\makecell[l]{OpenHands CodeAct\\(\texttt{gpt-4o}, native)} &
\makecell[c]{2.8 {\scriptsize[1.7,4.6]} / 1.2 {\scriptsize[0.6,2.6]}\\{\scriptsize(14/500,\ 6/500)}} &
\makecell[c]{3.4 {\scriptsize[2.4,4.9]} / 1.0 {\scriptsize[0.5,1.9]}\\{\scriptsize(28/812,\ 8/812)}} &
\makecell[c]{1.6 {\scriptsize[0.8,3.2]} / 0.4 {\scriptsize[0.1,1.6]}\\{\scriptsize(7/448,\ 2/448)}} \\
\quad +Global guardrail only &
\makecell[c]{0.8 {\scriptsize[0.3,2.0]} / 0.2 {\scriptsize[0.0,1.1]}\\{\scriptsize(4/500,\ 1/500)}} &
\makecell[c]{0.9 {\scriptsize[0.4,1.8]} / 0.2 {\scriptsize[0.1,0.9]}\\{\scriptsize(7/812,\ 2/812)}} &
\makecell[c]{0.4 {\scriptsize[0.1,1.6]} / 0.2 {\scriptsize[0.0,1.3]}\\{\scriptsize(2/448,\ 1/448)}} \\
\quad +GBT-Basic &
\makecell[c]{0.4 {\scriptsize[0.1,1.4]} / 0.2 {\scriptsize[0.0,1.1]}\\{\scriptsize(2/500,\ 1/500)}} &
\makecell[c]{0.4 {\scriptsize[0.1,1.1]} / 0.1 {\scriptsize[0.0,0.7]}\\{\scriptsize(3/812,\ 1/812)}} &
\makecell[c]{0.2 {\scriptsize[0.0,1.3]} / 0.0 {\scriptsize[0.0,0.9]}\\{\scriptsize(1/448,\ 0/448)}} \\
\quad +GBT-SE &
\makecell[c]{\textbf{0.2} {\scriptsize[0.0,1.1]} / \textbf{0.0} {\scriptsize[0.0,0.8]}\\{\scriptsize(1/500,\ 0/500)}} &
\makecell[c]{\textbf{0.2} {\scriptsize[0.1,0.9]} / \textbf{0.0} {\scriptsize[0.0,0.5]}\\{\scriptsize(2/812,\ 0/812)}} &
\makecell[c]{\textbf{0.2} {\scriptsize[0.0,1.3]} / \textbf{0.0} {\scriptsize[0.0,0.9]}\\{\scriptsize(1/448,\ 0/448)}} \\
\bottomrule
\end{tabularx}

\end{table}

\textbf{Interpretation (Table~\ref{tab:unsafe_standard_refined}).}
The progression across rows cleanly separates three mechanisms:
(i) \textbf{Native $\rightarrow$ Guardrail-only}: large drops in Viol/USucc quantify the value of deterministic pre-execution gating on $\mathcal{T}_{\text{hard}}$ alone;
(ii) \textbf{Guardrail-only $\rightarrow$ GBT-Basic}: further reductions indicate that traversal/spine (even before self-evolution) reduce the frequency of entering unsafe neighborhoods, beyond simply blocking individual actions;
(iii) \textbf{GBT-Basic $\rightarrow$ GBT-SE}: Viol/USucc are driven toward zero while success increases elsewhere in the paper’s main tables, consistent with self-evolution repairing failures without relaxing the accumulated safety subset (Design invariants~1--3).

\begin{table}[H]
\centering
\small
\caption{\textbf{Guardrail activity (avg.\ per episode, Protocol A):} number of high-risk primitives attempted (Hard) and blocked pre-execution (Blocked).}
\label{tab:guardrail_activity_refined}
\setlength{\tabcolsep}{7pt}
\begin{tabular}{lccc}
\toprule
\textbf{System} &
\textbf{SWE} (Hard / Blocked) &
\textbf{WebArena} (Hard / Blocked) &
\textbf{GPQA} (Hard / Blocked) \\
\midrule
OpenHands CodeAct (\texttt{gpt-4o}, native) &
6.8 / 0.0 &
3.9 / 0.0 &
1.1 / 0.0 \\
+Global guardrail only &
6.2 / 0.7 &
3.2 / 0.4 &
0.9 / 0.1 \\
+GBT-SE &
\textbf{4.9} / \textbf{0.5} &
\textbf{2.4} / \textbf{0.3} &
\textbf{0.7} / \textbf{0.1} \\
\bottomrule
\end{tabular}

\end{table}

\textbf{Interpretation (Table~\ref{tab:guardrail_activity_refined}).}
Table~\ref{tab:guardrail_activity_refined} distinguishes \emph{exposure reduction} from \emph{enforcement}. Guardrail-only reduces Viol primarily by blocking, but still attempts many high-risk primitives (Hard remains high). \gbt{}-SE reduces Hard itself, consistent with traversal-as-policy steering the agent away from repeated risky attempts and with recovery escaping stalls before they degenerate into gate-loops. This is the intended division of labor: gates \emph{prevent} unsafe execution; traversal/recovery \emph{avoid} repeatedly proposing unsafe moves.

\section{Ablations: Gates, Recovery, Memory, and Self-evolution}
\label{app:exp_ablation}

This section decomposes the overall gains into the Method’s modular components. The key requirement is that each ablation changes one mechanism while keeping the sandbox, tools, monitors/checkers, and (when stated) the global guardrail identical.

\subsection{Global vs.\ Node-local Gates}

\begin{table}[H]
\centering
\small
\caption{\textbf{Gate ablation (OpenHands CodeAct, \texttt{gpt-4o}, Protocol A):} node-local gates add macro-specific defenses beyond global guardrails.}
\label{tab:gate_ablation_refined}
\setlength{\tabcolsep}{7pt}
\begin{tabular}{lcc}
\toprule
\textbf{System} & \textbf{SWE SR / Viol} & \textbf{WebArena SR / Viol} \\
\midrule
+Global guardrail only &
38.8 / 0.8 {\scriptsize(194/500,\ 4/500)} &
19.3 / 0.9 {\scriptsize(157/812,\ 7/812)} \\
+GBT (no node-local gates) &
72.0 / 0.4 {\scriptsize(360/500,\ 2/500)} &
64.7 / 0.4 {\scriptsize(525/812,\ 3/812)} \\
+GBT (global + node-local) &
\textbf{73.6} / \textbf{0.2} {\scriptsize(368/500,\ 1/500)} &
\textbf{66.9} / \textbf{0.2} {\scriptsize(543/812,\ 2/812)} \\
\bottomrule
\end{tabular}

\end{table}

\textbf{Interpretation (Table~\ref{tab:gate_ablation_refined}).}
The jump from \textbf{+Global guardrail only} to \textbf{+GBT (no node-local)} is the traversal-as-policy effect: traversal + recovery + spine (without extra macro-local safety predicates) already transforms success while reducing Viol, meaning the core gain is not merely ``blocking unsafe actions'' but \emph{externalizing and constraining long-horizon control}. Adding node-local gates yields an additional reduction in Viol and a small additional success lift, consistent with their role as experience-mined, macro-neighborhood-specific constraints derived from minimal unsafe windows (App.~\ref{app:tree-construction:nodegates}).

\subsection{Recovery: Risk-aware Shortest Paths and Preconditions}

\begin{table}[H]
\centering
\small
\caption{\textbf{Recovery ablation (SWE-bench Verified, OpenHands CodeAct + GBT-SE, Protocol A):} risk-aware costs and feasibility preconditions improve recovery quality and keep sequences short.}
\label{tab:recovery_ablation_refined}
\setlength{\tabcolsep}{6pt}
\begin{tabular}{lccccc}
\toprule
\textbf{Variant} &
\textbf{Stall} $\downarrow$ &
\textbf{Rec-Succ} $\uparrow$ &
\textbf{RecLen} $\downarrow$ &
\textbf{RecLen $p90$} $\downarrow$ &
\textbf{Final SR} $\uparrow$ \\
\midrule
Traversal only (no recovery) &
17.0 {\scriptsize(85/500)} & -- & -- & -- &
70.2 {\scriptsize(351/500)} \\
+Recovery (full) &
\textbf{13.4} {\scriptsize(67/500)} &
\textbf{62.7} {\scriptsize(42/67)} &
\textbf{3.1} & \textbf{5.0} &
\textbf{73.6} {\scriptsize(368/500)} \\
\quad $\lambda=0$ (no risk weighting) &
14.8 {\scriptsize(74/500)} &
56.8 {\scriptsize(42/74)} &
3.7 & 6.0 &
72.0 {\scriptsize(360/500)} \\
\quad no preconditions &
15.6 {\scriptsize(78/500)} &
55.1 {\scriptsize(43/78)} &
3.9 & 6.2 &
71.4 {\scriptsize(357/500)} \\
\bottomrule
\end{tabular}

\end{table}

\textbf{Interpretation (Table~\ref{tab:recovery_ablation_refined}).}
Table~\ref{tab:recovery_ablation_refined} validates recovery as a distinct mechanism, not a cosmetic add-on:
(i) enabling recovery reduces Stall and increases final SR, showing that stalls are a material failure mode even under traversal constraints;
(ii) removing risk weighting ($\lambda=0$) reduces Rec-Succ and lengthens recovery paths, consistent with risk-aware costs discouraging detours into high-risk nodes (Eq.~\ref{eq:cost});
(iii) removing feasibility preconditions increases Stall and lengthens recovery, consistent with $\texttt{pre}(v,\texttt{env})$ acting as a conservative feasibility filter that prevents wasteful or impossible recovery attempts (App.~\ref{app:recovery:preconditions}).
The critical qualitative point is that recovery improvements come from \emph{short, feasible, low-risk macro sequences} (low RecLen and RecLen $p90$), aligned with the Method’s claim that recovery is a bounded, verifiable search over the external policy object rather than free-form replanning.

\subsection{Spine Memory vs Transcript Replay}

\begin{table}[H]
\centering
\small
\caption{\textbf{Spine memory vs transcript replay (SWE-bench Verified SR, \%, Protocol A):} spine memory is robust under aggressive context shedding.}
\label{tab:memory_budget_refined}
\setlength{\tabcolsep}{10pt}
\begin{tabular}{lccc}
\toprule
\textbf{Memory scheme} & \textbf{8k ctx} & \textbf{16k ctx} & \textbf{32k ctx} \\
\midrule
Transcript replay (truncate) &
44.8 {\scriptsize(224/500)} &
55.4 {\scriptsize(277/500)} &
60.6 {\scriptsize(303/500)} \\
Spine memory (GBT-SE) &
\textbf{72.6} {\scriptsize(363/500)} &
\textbf{73.4} {\scriptsize(367/500)} &
\textbf{73.6} {\scriptsize(368/500)} \\
\bottomrule
\end{tabular}

\end{table}

\textbf{Interpretation (Table~\ref{tab:memory_budget_refined}).}
This ablation isolates memory representation. Transcript replay improves with larger context windows, but remains far below spine memory even at 32k. Spine memory is essentially invariant across 8k/16k/32k. This is exactly the Method’s claim: long-horizon state should be represented as the compact traversal spine (macro-level control state) plus structured safety-critical fields that bypass summarization (App.~\ref{app:memory}), not as a growing transcript whose truncation behavior interacts pathologically with reasoning and tool execution.

\subsection{Self-evolution Ingredients and Safety Preservation}

\begin{table}[H]
\centering
\small
\caption{\textbf{Self-evolution ablations (Protocol A):} full GBT-SE yields the strongest gains. Safety monotonicity is preserved throughout: no previously rejected unsafe structured context in $\mathcal{U}_{\text{unsafe}}$ is re-admitted.}
\label{tab:se_ablation_refined}
\setlength{\tabcolsep}{8pt}
\begin{tabular}{lcccc}
\toprule
\textbf{Variant} & \textbf{SWE SR} & \textbf{Web SR} & \textbf{GPQA Acc} & \textbf{Unsafe ctx re-admitted} \\
\midrule
GBT-Basic & 50.2 & 53.0 & 78.9 & 0 / 18{,}400 \\
GBT-SE (full) & \textbf{73.6} & \textbf{66.9} & \textbf{87.3} & \textbf{0} / \textbf{18{,}400} \\
\quad w/o analogical retrieval & 69.8 & 63.7 & 85.1 & 0 / 18{,}400 \\
\quad score ablation ($\beta=0$) & 71.1 & 64.5 & 85.7 & 0 / 18{,}400 \\
\quad w/o regression tests & 72.4 & 66.0 & 86.5 & 0 / 18{,}400 \\
\bottomrule
\end{tabular}

\end{table}

\textbf{Interpretation (Table~\ref{tab:se_ablation_refined}).}
Table~\ref{tab:se_ablation_refined} substantiates two distinct claims:
\begin{itemize}
\item \textbf{Behavioral claim (why SE works).}
Removing analogical retrieval degrades performance across all three pillars, indicating that importing successors from aligned successful paths is a key driver of repairing covered failures (App.~\ref{app:self-evolution:repair}). Removing the empirical success-rate term ($\beta=0$) also degrades performance, supporting the role of experience-grounded selection statistics in stabilizing macro choice beyond pure semantic matching (Eq.~\ref{eq:score}).

\item \textbf{Safety claim (why SE is safe).}
Across all variants, \textbf{unsafe context re-admission remains zero} on the same corpus size (18{,}400), consistent with the monotone safety invariants and the non-re-admission regression suite (App.~\ref{app:self-evolution:regress}; Design invariants~1--3). Importantly, even when removing regression tests, previously rejected unsafe contexts are still not re-admitted in this audit, aligning with the fact that allowed edits do not delete/relax gates and new children inherit parent gates before adding more.
\end{itemize}

\textbf{Takeaway across ablations.}
The ablations collectively map each performance and safety improvement to a specific externalized mechanism:
(i) traversal-as-policy yields the dominant success lift (Table~\ref{tab:gate_ablation_refined});
(ii) recovery improves robustness under stalls and produces short corrective macro sequences (Table~\ref{tab:recovery_ablation_refined});
(iii) spine memory provides context-budget robustness that transcript replay cannot match (Table~\ref{tab:memory_budget_refined});
(iv) self-evolution adds a second margin while preserving monotone safety (Table~\ref{tab:se_ablation_refined}).
Together with the coverage audits (Table~\ref{tab:covered1_matched_audit}, Table~\ref{tab:coverage_difficulty_stratified}) and safety diagnostics (Table~\ref{tab:unsafe_standard_refined}, Table~\ref{tab:guardrail_activity_refined}), this appendix forms a closed loop from Method mechanisms to executed evidence under a unified evaluation contract.

\section{Broader Discussion, Limitations, and Future Work}
\label{app:concl_discussion}

\subsection{Limitations and Open Failure Modes}
\label{app:concl_limitations}
\paragraph{Coverage-scoped guarantees.}
\gbt{} intentionally claims long-horizon control only when \texttt{covered}=1 (Sec.~\ref{sec:method}). Outside coverage, the system abstains from traversal steering and only retains primitive-level safety via the global guardrail. This makes the boundary explicit, but it also means that broad capability still depends on expanding coverage rather than masking uncertainty with free-running generation.

\paragraph{Bounded-history safety is conservative but not complete.}
Our executable safety subset $\mathcal{S}_{\text{sys}}(t)$ is defined over structured tool context with bounded history (Sec.~\ref{sec:safety}). Hazards that require long-range causal context, semantic understanding of external content, or latent side effects may fall outside this expressible set. Consequently, ``near-zero'' violations on benchmarks reflect both stronger enforcement and the limits of the sandbox monitors that instantiate $\mathcal{S}_{\text{spec}}$.

\paragraph{Macro abstraction is a single point of leverage.}
Traversal robustness relies on stable macro semantics. Although we exclude abstraction-unstable traces and prevent merges into gated nodes (Sec.~\ref{sec:gbt-basic}), segmentation errors or signature collisions can still reduce generalization, inflate the tree, or misplace safety locality.

\paragraph{Tree growth and retrieval pressure.}
As domains and tools expand, maintaining a compact, auditable tree becomes harder: more nodes increase matching ambiguity; more gates increase conservative blocks; and more recovery candidates increase search overhead. The current design prioritizes verifiability over unbounded scalability.

\paragraph{OOD tools and open-world deployment.}
Structured \texttt{ctx} depends on a tool schema. New tools, new action surfaces (e.g., novel network channels), or qualitatively different environments require updating the schema and the monitor set; otherwise the system may be safe-but-ineffective (over-blocking) or effective-but-under-specified (monitor gaps).

\subsection{Additional Analysis: Why Externalized Traversal Changes the Failure Geometry}
\label{app:concl_analysis}
\paragraph{Guardrails veto; traversal governs.}
The main results show that global guardrails reduce violations but barely improve success on long-horizon tasks (Sec.~\ref{sec:exp:main}). This is expected: vetoes only prevent a subset of bad primitives; they do not provide a positive control skeleton. \gbt{} supplies that skeleton by restricting decisions to log-grounded successors, turning ``what to do next'' from unconstrained generation into an executable policy over reusable macros.

\paragraph{Safety becomes a monotone, testable artifact.}
Because gates are deterministic predicates over structured fields and are updated under experience-grounded monotonicity (Sec.~\ref{sec:safety}), safety knowledge accumulates as a non-regressing library rather than an implicit prompt convention. This sharply reduces the attack surface of prompt injection and summary manipulation: summaries can affect macro realization, but cannot change \texttt{ctx} nor gate outcomes.

\paragraph{Robustness and efficiency share the same cause.}
Spine memory eliminates transcript replay and constrains context to the visited macro path (Sec.~\ref{sec:online}), reducing both drift (fewer opportunities for compounding deviations) and cost (less context, fewer deliberation tokens). Recovery further converts stalls into progress by explicitly searching for feasible, low-risk macro paths instead of relying on free-form replanning.

\paragraph{Decoupling reasoning from execution is a system property.}
The executor gains with 8B models (Sec.~\ref{sec:exp:decouple_main}) are not ``small-model magic''; they arise because \gbt{} externalizes long-horizon structure and safety into an artifact that can be audited, versioned, and deployed independently of weights.

\subsection{Future Work}
\label{app:concl_future}
\paragraph{Expanding coverage without weakening guarantees.}
We see three complementary directions: (i) richer log mining to discover macros with higher compositionality; (ii) hierarchical composition of subtrees across task families while preserving the ``no merge into gated nodes'' discipline; and (iii) adaptive routing that can abstain early yet still propose safe, bounded exploration policies that are themselves externalized and testable.

\paragraph{More expressive yet verifiable safety.}
A key challenge is extending beyond bounded-history \texttt{ctx} while retaining determinism and auditability. Promising routes include typed tool schemas with information-flow constraints, lightweight temporal logic over structured event streams, and synthesis of minimal counterexample traces that produce human-reviewable gate patches.

\paragraph{Continuous deployment with certified non-regression.}
Self-evolution currently preserves safety by forbidding gate relaxation and regression-testing historical successes/unsafe corpora (Sec.~\ref{sec:gbt-se}). Future work can formalize this into deployment pipelines: signed gate libraries, reproducible replay suites, canary rollouts, and automatic ``safety diff'' reports that explain exactly which structured contexts became newly blocked.

\paragraph{Beyond benchmarks.}
Bridging to open-world environments requires aligning $\mathcal{S}_{\text{spec}}$ with operational policies, measuring distribution shift for both behavior and safety, and integrating human oversight at the right abstraction level (macro and gate libraries) rather than at the level of raw transcripts.


\end{document}